\documentclass[9pt,twocolumn,twoside]{osajnl}
\usepackage{stfloats}
\usepackage{float}
\usepackage{subfigure}
\usepackage{graphicx}
\usepackage{amsmath}
\usepackage{amssymb}
\usepackage{multirow}
\usepackage{setspace}
\newcommand{\bb}[1]{{\mathbf{#1}}}
\newcommand{\nn}{\nonumber}
\usepackage{color}

\journal{josaa} 

\setboolean{shortarticle}{false} 

\title{Two-dimensional nonseparable discrete linear canonical transform based on CM-CC-CM-CC decomposition}

\author[1,*]{Soo-Chang Pei}
\author[2]{Shih-Gu Huang}

\affil[1]{Department of Electrical Engineering \& Graduate Institute of Communication Engineering, National Taiwan University, Taipei 10617, Taiwan}
\affil[2]{Graduate Institute of Communication Engineering, National Taiwan University, Taipei 10617, Taiwan}

\affil[*]{Corresponding author: peisc@ntu.edu.tw}

\dates{Compiled November 22, 2015}

\ociscodes{(070.2575) Fractional Fourier transforms; (070.2580) Paraxial wave optics; (070.2590) ABCD transforms;  (070.0070) Fourier optics and signal processing;
(080.2730) Matrix methods in paraxial optics.}

\doi{\url{http://dx.doi.org/10.1364/ao.XX.XXXXXX}}

\begin{abstract}
As a generalization of the two-dimensional Fourier transform (2D FT) and 2D fractional Fourier transform, the 2D nonseparable linear canonical transform (2D NsLCT) is useful in optics, signal and image processing.
To reduce the digital implementation complexity of  the 2D NsLCT, some previous works decomposed the 2D NsLCT into several low-complexity operations, including 2D FT, 2D chirp multiplication (2D CM) and 2D affine transformations.
However, 2D affine transformations will introduce interpolation error.
In this paper, we propose a new decomposition called CM-CC-CM-CC decomposition, which decomposes the 2D NsLCT into two 2D CMs and two 2D chirp convolutions (2D CCs).
No 2D affine transforms are involved.
Simulation results show that the proposed methods have higher accuracy, lower computational complexity and smaller error in the additivity property compared with the  previous works.
Plus, the proposed methods have perfect reversibility property that one can reconstruct the input signal/image losslessly from the output.
\normalfont {\small{© 2015 Optical Society of America. One print or electronic copy may be made for personal use only. Systematic reproduction and distribution, duplication of any material in this paper for a fee or for commercial purposes, or modifications of the content of this paper are prohibited.}}
\end{abstract}

\setboolean{displaycopyright}{false}

\begin{document}

\maketitle
\thispagestyle{fancy}
\ifthenelse{\boolean{shortarticle}}{\abscontent}{}

\vspace*{-32pt}
\section{Introduction}\label{sec:Intro}
Linear canonical transform (LCT), first introduced in \cite{collins1970lens,moshinsky1971linear}, is a generalization of the fractional Fourier transform (FRFT).
It unifies a variety of transforms such as Fourier transform (FT),  FRFT and Fresnel transform.
It has four parameters with three degrees of freedom, and thus more important and useful in
optics
\cite{nazarathy1982first,bastiaans1989propagation,ozaktas2001fractional} and many signal processing applications including filter design, radar  system  analysis, signal synthesis, phase  reconstruction, time-frequency analysis, pattern  recognition, encryption and modulation \cite{barshan1997optimal,pei2001relations,bastiaans2003phase,hennelly2005optical,sharma2006signal,pei2013reversible}.
To extend the 1D LCT to two dimensions $(x,y)$, an easy and straightforward approach is performing two independent 1D LCTs in the two transverse directions $x$ and $y$, respectively.
Since the two-dimensional (2D) kernel can be separated, this 2D transform is called 2D separable LCT (2D SLCT) \cite{sahin1998optical}.
The 2D SLCT can produce affine transformations in the $(x,\omega_x)$ and $(y,\omega_y)$ planes, where $\omega_x$ and $\omega_y$ are the spatial-frequency coordinates with respect to $x$ and $y$.
Since the 1D LCT has three degrees of freedom, the 2D SLCT has six degrees of freedom. 

A further generalization of the 2D SLCT is the 2D nonseparable LCT (2D NsLCT) \cite{folland1989harmonic,pei2001two,alieva2005alternative}, named after its nonseparable 2D kernel.
The 2D NsLCT provides four more (i.e. ten) degrees of freedom to represent all transformations not only  in  $(x,\omega_x)$ and $(y,\omega_y)$ planes but also in $(x,y)$, $(x,\omega_y)$, $(\omega_x,y)$  and $(\omega_x,\omega_y)$ planes.
The 1D/2D FT, FRFT and Fresnel transform, as well as the 1D LCT, 2D SLCT and gyrator transform \cite{rodrigo2007gyrator} are all its special cases.
All the applications of these special cases in optics,  signal processing and digital image processing can be extended and become more flexible by the 2D NsLCT.
For example, in \cite{pei2001two}, the authors show that the noise with nonseparable term cannot be removed clearly by the 2D SLCT filter but can be by the 2D NsLCT filter.
In optical system analysis, the 2D NsLCT is more effective when analyzing systems containing quadratic phase components misaligned in both $x$ and $y$ axes \cite{bastiaans2007classification,ding2011eigenfunctions}.

Consider an affine transformation in the space-spatial-frequency plane:
\begin{align}\label{eq:Intro04}
\begin{bmatrix}
u\\
v\\
{{\omega _u}}\\
{{\omega _v}}
\end{bmatrix}
= 
\begin{bmatrix}
\bb A & \bb B\\
\bb C & \bb D
\end{bmatrix}
\begin{bmatrix}
x\\
y\\
{{\omega _x}}\\
{{\omega _y}}
\end{bmatrix}
= \begin{bmatrix}
a_{11}&a_{12}&b_{11}&b_{12}\\
a_{21}&a_{22}&b_{21}&b_{22}\\
c_{11}&c_{12}&d_{11}&d_{12}\\
c_{21}&c_{22}&d_{21}&d_{22}
\end{bmatrix}
\begin{bmatrix}
x\\
y\\
{{\omega _x}}\\
{{\omega _y}}
\end{bmatrix}.
\end{align}
where $(\omega_x,\omega_y)$ and $(\omega_u,\omega_v)$ are the spatial-frequency coordinates with respect to $(x,y)$ and $(u,v)$, respectively.
The $4\times4$ 
transformation matrix in (\ref{eq:Intro04}) is called ABCD matrix and also denoted by $(\bb A,\bb B;\bb C,\bb D)$ in this paper.
Assume $\bb z=[x,y]^T$ and $\bb z'=[u,v]^T$.
The 2D NsLCT that can result in the space-spatial-frequency transformation in (\ref{eq:Intro04}) is given by \cite{folland1989harmonic,pei2001two,alieva2005alternative}
\begin{align}\label{eq:Intro08}
G(\bb z')&\!= {\cal O}_{\textmd{NsLCT}}^{(\bb A,\bb B;\bb C,\bb D)}\!\{g(\bb z)\}\!=\!\frac{1}{{2\pi\! \sqrt { - \det ({\bf{B}})} }}\!\int\! {\exp\! \left[ {\frac{j}{2}\!\left( {{{{\bf{z'}}}^T}{\bf{D}}{{\bf{B}}^{ - 1}}{\bf{z'}}} \right.} \right.}\nn\\
&\qquad\qquad\qquad\quad \left. {\left. {\frac{{}}{{}} - 2{{\bf{z}}^T}{{\bf{B}}^{ - 1}}{\bf{z'}} + {{\bf{z}}^T}{{\bf{B}}^{ - 1}}{\bf{Az}}} \right)} \right]g(\bb z)d\bb z,\!
\end{align}
where
$g(\bb z)=g(x,y)$ and $G(\bb z')=G(u,v)$ are the 2D input and output signals, respectively.
The ABCD matrix $(\bb A, \bb B; \bb C, \bb D)$ for a valid 2D NsLCT should satisfy
\begin{align}\label{eq:Intro12}
{\bf{A}}{{\bf{B}}^T} = {\bf{B}}{{\bf{A}}^T},\ \ \ {\bf{C}}{{\bf{D}}^T} = {\bf{D}}{{\bf{C}}^T}, \ \ \ {\bf{A}}{{\bf{D}}^T}-{\bf{B}}{{\bf{C}}^T}=\bb I,
\end{align}
or equivalently
\begin{align}\label{eq:Intro16}
{{\bf{A}}^T}{\bf{C}} = {{\bf{C}}^T}{\bf{A}},\ \ \ {{\bf{B}}^T}{\bf{D}} = {{\bf{D}}^T}{\bf{B}},\ \ \
{{\bf{A}}^T}{\bf{D}}-{{\bf{C}}^T}{\bf{B}}=\bb I.
\end{align}
Either (\ref{eq:Intro12}) or (\ref{eq:Intro16}) leads to six linear equations, i.e. six constraints.
Although there are a total of $16$ parameters in 
$\bb A$, $\bb B$, $\bb C$ and $\bb D$, the number of degrees of freedom is 10 due to the six constraints.
It is obvious that the definition in (\ref{eq:Intro08}) is valid only when $\det(\bb B)\neq0$, i.e. $\bb B$ is invertible.
When $\bb B=\bb 0$, the 2D NsLCT reduces into a 2D affine transformation multiplied by a 2D chirp function:
\begin{align}\label{eq:Intro20}
G(\bb z')&= \sqrt {\det ({\bf{D}})} \exp \left( {\frac{j}{2}{{{\bf{z'}}}^T}{\bf{C}}{{\bf{D}}^T}{\bf{z'}}} \right)g\left( {{{\bf{D}}^T}{\bf{z'}}} \right).
\end{align}
When $\det(\bb B)=0$ but $\bb B\neq0$, the definition of 2D NsLCT is subdivided into several different cases.
One can refer to \cite{pei2001two} for a detailed details.
%
The digital implementation of the 1D LCT has been widely studied in many papers such as \cite{pei2000closed,hennelly2005fast,ozaktas2006efficient,kelly2006analytical,koc2008digital,healy2010reevaluation,healy2010fast,pei2011discrete}.
The digital implementation of the 2D SLCT can be easily realized by performing any of these 1D implementation techniques twice, one in the $x$ direction and one in the $y$ direction.
However, there are less works regarding the digital implementation of the 2D NsLCT \cite{kocc2010fast,ding2012improved,zhao2013unitary,zhao2014sampling,zhao2014two}.
Zhao \emph{et al.}'s works \cite{zhao2013unitary,zhao2014sampling,zhao2014two} mainly focus on the sampling of the 2D NsLCT to ensure unitary property.
Ko{\c{c}} \emph{et al.}'s work \cite{kocc2010fast} and Ding \emph{et al.}'s work \cite{ding2012improved} focus on the development of digital implementation algorithms to improve complexity and accuracy.

To develop the 2D NsDLCT, the simplest method is to discretize the 2D NsLCT in (\ref{eq:Intro20}) by sampling and summation:
\begin{align}\label{eq:Intro32}
G[p,q]&={\cal O}_{\textmd{Direct}}^{(\bb A,\bb B;\bb C,\bb D)}\!\{g[m,n]\}=\frac{\Delta_x\Delta_y}{{2\pi\! \sqrt { - \det ({\bf{B}})} }}\sum\limits_m \sum\limits_n \exp\! \left[ \frac{j}{2}  \right.\nn\\
&\quad\left. \frac{}{}\cdot\!\left(\! {{{{\bf{r'}}}^T}{\bf{D}}{{\bf{B}}^{ - 1}}{\bf{r'}}} - 2{{\bf{r}}^T}{{\bf{B}}^{ - 1}}{\bf{r'}} + {{\bf{r}}^T}{{\bf{B}}^{ - 1}}{\bf{Ar}} \right) \right]g[m,n],
\end{align}
where $\bb r=[m\Delta_x,n\Delta_y]^T$ and $\bb r'=[p\Delta_u,q\Delta_v]^T$
are the input and output sampling points, respectively.
This 2D NsDLCT is named direct method in this paper.
The direct method is very inefficient.
In order to reduce computational complexity, the 2D NsLCT is decomposed into several simpler 2D operations, and it follows that one can develop a low-complexity 2D NsDLCT by connecting several low-complexity 2D discrete operations.
Decomposition methods have been widely used in the digital implementation of 1D/2D FRFT, 1D LCT, gyrator transform, etc.
In \cite{kocc2010fast}, the authors decomposed the 2D NsLCT into one 2D chirp multiplication (2D CM), one 2D FRFT and two 2D affine transformations.
However, 2D affine transformations will introduce interpolation error in the discrete case.
According, in \cite{ding2012improved}, another decomposition that involves two 2D CMs, one 2D FT and only one 2D affine transformation is proposed.

In this paper, the accuracy is further improved by decomposing the 2D NsLCT into two 2D CMs and two 2D chirp convolutions (2D CCs), called CM-CC-CM-CC decomposition.
More precisely, only 2D CMs, 2D FTs and 2D inverse FTs (2D IFTs) are used.
No 2D affine transformations are involved.
Based on this decomposition, two types of 2D NsDLCT are proposed.
Compared to each other, one has higher accuracy while the other one has lower complexity.
All the proposed 2D NsDLCTs have lower complexity and higher accuracy than the previous works \cite{kocc2010fast,ding2012improved}.
Plus, the proposed methods have lower error in additivity property, which is a useful property in applications such as filtering and encryption/decryption.
Besides, another decomposition called CM-CC-CM-CC decomposition is introduced such that the proposed methods have perfect reversibility property.
That is, the input discrete signal/image can be perfectly recovered from the output of the proposed 2D NsDLCTs.

%
%
%
%
%
%

\section{Basic 2D Discrete Operations}\label{sec:Basic}
The 2D nonseparable discrete LCTs (NsDLCTs) proposed in this paper will be compared with Ko{\c{c}}'s method \cite{kocc2010fast} and  Ding's method \cite{ding2012improved}.
In this section, some basic 2D discrete operations used in these 2D NsDLCTs are
introduced.
The computational complexity
of these operations is also analyzed.
Denote $G(u,v)$ as the output of a 2D continuous operation with $g(x,y)$ being the input.
Given $N\times N$ sampled input signal $g[m,n]\buildrel \Delta \over =g(m\Delta_x,n\Delta_y)$,
the corresponding 2D discrete operation is designed to generate $N\times N$ output $G[p,q]$ that can approximate $G(p\Delta_u,q\Delta_v)$.

\subsection{2D Discrete Chirp Multiplication (Discrete CM)}\label{subsec:BasicCM}
Consider that the ABCD matrix is given by
\begin{align}\label{eq:BasicCM04}
\begin{bmatrix}
{\bf{A}}&{\bf{B}}\\
{\bf{C}}&{\bf{D}}
\end{bmatrix} = \begin{bmatrix}
{\bf{I}}&{\bf{0}}\\
{\bf{C}}&{\bf{I}}
\end{bmatrix},\quad\textmd{where}\quad
\bb C=\begin{bmatrix}
c_{11}&c_{12}\\
c_{12}&c_{22}
\end{bmatrix}
\end{align}
is symmetric, i.e.  $\bb C=\bb C^T$, because of the constraints in (\ref{eq:Intro12}) or (\ref{eq:Intro16}). 
Thus, this ABCD matrix has only three degrees of freedom.
Since $\bb B=\bb 0$, the definition in (\ref{eq:Intro20}) is used, and
the 2D NsLCT becomes 
the multiplication with a 2D chirp function, called 2D chirp multiplication (CM) for short:
\begin{align}\label{eq:BasicCM08}
G(u,v)= {\cal O}_{\textmd{NsLCT}}^{(\bb I, \bb0;\bb C, \bb I)}\{g(x,y)\}\!={e^{\frac{j}{2}\left( c_{11}{u^2} + 2c_{12}uv + c_{22}{v^2} \right)}}g(u,v).
\end{align}
Sampling (\ref{eq:BasicCM08}) with sampling intervals $\Delta_u=\Delta_x$ and $\Delta_v=\Delta_y$, the 2D discrete CM, denoted by ${\cal O}_{\textmd{CM}}^{\bb C}$, is given by
\begin{align}\label{eq:BasicCM12}
G[p,q]&= {\cal O}_{\textmd{CM}}^{\bb C}\{g[m,n]\}\buildrel \Delta \over ={e^{\frac{j}{2}\left( c_{11}{p^2}\Delta_u^2 + 2c_{12}pq\Delta_u\Delta_v + c_{22}{q^2}\Delta_v^2 \right)}}g[p,q].
\end{align}
Supposing the exponential kernel function in (\ref{eq:BasicCM12}) can be precomputed and stored in memory,
only one $N\times N$ pointwise product that involves $N^2$ complex multiplications is required.

\subsection{2D Discrete Fourier Transform (DFT) and inverse DFT (IDFT)}\label{subsec:BasicDFT}
In (\ref{eq:Intro08}), if ABCD matrix is given by
\begin{align}\label{eq:BasicDFT04}
\begin{bmatrix}
{\bf{A}}&{\bf{B}}\\
{\bf{C}}&{\bf{D}}
\end{bmatrix} = \begin{bmatrix}
{\bf{0}}&{\bf{I}}\\
{\bf{-I}}&{\bf{0}}
\end{bmatrix} \quad \textmd{or} \quad \begin{bmatrix}
{\bf{0}}&{\bf{-I}}\\
{\bf{I}}&{\bf{0}}
\end{bmatrix},
\end{align}
the 2D NsLCT reduces to the 2D Fourier transform (FT) or the 2D inverse FT (IFT) multiplied by constant phase $1/j$:
\begin{align}\label{eq:BasicDFT08}
G(u,v)= \frac{1}{{j2\pi }}\!\int\limits_{ - \infty }^\infty  {\int\limits_{ - \infty }^\infty  \!{e ^ { \mp jux \mp jvy}g(x,y)dxdy} }.
\end{align}
If $\Delta_u\Delta_x=\Delta_v\Delta_y=2\pi/N$, the discrete version of (\ref{eq:BasicDFT08}),
denoted by ${\cal F}_{x,y}$ or ${\cal F}^{-1}_{x,y}$, are simply the 2D DFT or 2D IDFT multiplied by some constant:
\begin{align}\label{eq:BasicDFT16}
G[p,q]&\!=\!{\cal F}_{x,y}\!\{g[m,n]\}\!  \buildrel \Delta \over = \frac{{{\Delta _x}{\Delta _y}}}{{j2\pi }}\sum\limits_m^{} {\sum\limits_n^{} {{e^{ - j\frac{{2\pi }}{N}(pm + qn)}}g[m,n]} },\\
G[p,q]&\!=\!{\cal F}_{x,y}^{-1}\!\{g[m,n]\} \!\buildrel \Delta \over =\frac{{{\Delta _x}{\Delta _y}}}{{j2\pi }}\sum\limits_m^{} {\sum\limits_n^{} {{e^{ + j\frac{{2\pi }}{N}(pm + qn)}}g[m,n]} }. \label{eq:BasicDFT20}
\end{align}
The 2D DFT and IDFT can be implemented by 2D FFT with $\frac{{{N^2}}}{2}{{\log }_2}{N^2}$ complex multiplications.
Zero-padding the input
signal to size $N'\times N'$, 
where $N'>N$, can 
reduce the
output sampling intervals to $\Delta_u=\frac{2\pi}{N'\Delta_x}$ and $\Delta_v=\frac{2\pi}{N'\Delta_y}$, but the cost is higher computational complexity.

\subsection{2D Discrete Chirp Convolution (Discrete CC)}\label{subsec:BasicCC}
Suppose the ABCD matrix is of the following form
\begin{align}\label{eq:BasicCC04}
\begin{bmatrix}
{\bf{A}}&{\bf{B}}\\
{\bf{C}}&{\bf{D}}
\end{bmatrix} = \begin{bmatrix}
{\bf{I}}&{\bf{B}}\\
{\bf{0}}&{\bf{I}}
\end{bmatrix},
\quad\textmd{where}\quad
\bb B=\begin{bmatrix}
b_{11}&b_{12}\\
b_{12}&b_{22}
\end{bmatrix}.
\end{align}
$\bb B$ is symmetric because of the constraints in (\ref{eq:Intro12}) or (\ref{eq:Intro16}) and thus has only three degrees of freedom.
In (\ref{eq:Intro08}), $(\bb I, \bb B;\bb 0, \bb I)$ leads to
\begin{align}\label{eq:BasicCC08}
&G(u,v)={\cal O}_{\textmd{NsLCT}}^{(\bb I, \bb B;\bb 0, \bb I)}\{g(x,y)\}=\frac{1}{{2\pi \sqrt { - \det ({\bf{B}})} }}\nn\\
&\!\!\cdot\!\int\limits_{ - \infty }^\infty\!\int\limits_{ - \infty }^\infty \!\!  {e^{\frac{j}{{2\det ({\bf{B}})}}\left[ {{b_{22}}{{(u - x)}^2} - 2{b_{12}}(u - x)(v - y) + {b_{11}}{{(v - y)}^2}} \right]}}\!g(x,y)dxdy,\!
\end{align}
which is a 2D convolution with a 2D chirp function and called 2D chirp convolution (CC) for short.
In the discrete case, directly calculating the sampled version of (\ref{eq:BasicCC08}) by summation 
leads to computational complexity up to $O(N^4)$.
Fortunately, the ABCD matrix in (\ref{eq:BasicCC04}) can be decomposed as
\begin{align}\label{eq:BasicCC12}
\begin{bmatrix}
{\bf{I}}&{\bf{B}}\\
{\bf{0}}&{\bf{I}}
\end{bmatrix} = \underbrace{\begin{bmatrix}
{\bf{0}}&{ - {\bf{I}}}\\
{\bf{I}}&{\bf{0}}
\end{bmatrix}}_{\textmd{2D IFT}}
\underbrace{\begin{bmatrix}
{\bf{I}}&{\bf{0}}\\
{ - {\bf{B}}}&{\bf{I}}
\end{bmatrix}}_{\textmd{2D CM}}
\underbrace{\begin{bmatrix}
{\bf{0}}&{\bf{I}}\\
{ - {\bf{I}}}&{\bf{0}}
\end{bmatrix}}_{\textmd{2D FT}}.
\end{align}
Taking the benefit of the additivity property of 2D NsLCT, the above equality implies that the 2D CC can be 
alternatively implemented by
one 2D IFT, one 2D CM and one 2D FT, i.e.
\begin{align}\label{eq:BasicCC16}
{\cal O}_{\textmd{NsLCT}}^{(\bb I, \bb B;\bb 0, \bb I)}=
{\cal O}_{\textmd{NsLCT}}^{(\bb 0, -\bb I;\bb I, \bb 0)}\,
{\cal O}_{\textmd{NsLCT}}^{(\bb I, \bb 0;-\bb B, \bb I)}\,
{\cal O}_{\textmd{NsLCT}}^{(\bb 0, \bb I;-\bb I, \bb 0)}.
\end{align}
Therefore, the 2D discrete CC with chirp matrix $\bb B$,
denoted by 
${\cal O}_{\textmd{CC}}^{\bb B}$, is defined as a cascade of one 2D IDFT, one 2D discrete CM
with chirp matrix $-\bb B$, and one 2D DFT:
\begin{align}\label{eq:BasicCC20}
{\cal O}_{\textmd{CC}}^{\bb B}\buildrel \Delta \over = {\cal F}^{-1}_{x,y}\:
{\cal O}_{\textmd{CM}}^{-\bb B}\:
{\cal F}_{x,y}.
\end{align}
Two 2D FFTs and one pointwise product are used and  totally require  $N^2{{\log }_2}{N^2}+N^2$ complex multiplications.

\subsection{2D Discrete Fractional Fourier Transform (DFRFT)}\label{subsec:BasicFRFT}
The 2D FRFT with two parameters $\alpha$ and $\beta$ is a special  case of the 2D NsLCT.
If $\bb A$, $\bb B$, $\bb C$ and $\bb D$   are given by
\begin{align}\label{eq:BasicFRFT04}
{\bf{A}} = {\bf{D}} = \begin{bmatrix}
{\cos \alpha }&0\\
0&{\cos \beta }
\end{bmatrix}, \ \
{\bf{B}} =  - {\bf{C}} = \begin{bmatrix}
{ \sin \alpha }&0\\
0&{ \sin \beta }
\end{bmatrix},
\end{align}
the 2D NsLCT 
reduces to 
the 2D FRFT \cite{sahin1995optical} with some constant phase difference:
\begin{align}\label{eq:BasicFRFT08}
&G(u,v)= {\cal O}_{\textmd{NsLCT}}^{(\bb A, \bb B;-\bb B,\bb A)}\{g(x,y)\}\nn\\
&=\frac{1}{{2\pi \sqrt { - \sin \alpha \sin \beta } }}\int\limits_{ - \infty }^\infty   \int\limits_{ - \infty }^\infty   {K_\alpha }(u,x){K_\beta }(v,y)g(x,y)dxdy,
\end{align}
where 
${K_\alpha}$ and ${K_\beta}$ are 1D FRFT kernels \cite{namias1980fractional,almeida1994fractional} with fractioanl angles $\alpha$ and $\beta$, respectively:
\begin{align}\label{eq:BasicFRFT12}
{K_\alpha }(u,x) = \exp \left( {j\frac{\cot \alpha}{2}   {u^2} - j\csc \alpha   ux + j\frac{\cot \alpha}{2}   {x^2}} \right).
\end{align}
Obviously, the 2D FRFT 
is separable and 
can be implemented by two 1D FRFTs in two transverse directions, $x$ and $y$.

There are a variety of implementation algorithms for 1D FRFT, and a review of some of them is given in \cite{pei2000closed,sejdic2011fractional}.
Here, we introduce the algorithm used in Ko{\c{c}}'s method \cite{kocc2010fast}.
If 
$\Delta_u\Delta_x=\frac{2\pi|\sin\alpha|}{N}$,
the sampled version of the kernel in (\ref{eq:BasicFRFT12}) is given by
\begin{align}\label{eq:BasicFRFT16}
{K_\alpha }[p,m] = \exp \!\left( \frac{j\cot \alpha}{2}   {p^2}\Delta_u^2 \mp j \frac{2\pi}{N} pm + \frac{j\cot \alpha}{2}    {m^2}\Delta_x^2 \right).
\end{align}
For the minus-plus sign $\mp$ in the above kernel,  minus is used
when $\sin\alpha>0$ while plus is used when $\sin\alpha<0$.
(\ref{eq:BasicFRFT16}) shows that
the 1D DFRFT can be implemented by two discrete CMs and one DFT/IDFT.

Once the 1D DFRFT is developed, the 2D DFRFT can be commutated by
two separate 1D DFRFTs in two transverse directions, $m$ and $n$:
\begin{align}\label{eq:BasicFRFT18}
G[p,q]&= {\cal F}^{\:\alpha,\beta}_{x,y}\{g[m,n]\}\nn\\
&\buildrel \Delta \over =\frac{\Delta_x\Delta_y}{{2\pi \sqrt { - \sin \alpha \sin \beta } }}\sum\limits_m^{} \sum\limits_n^{}   {K_\alpha }[p,m]{K_\beta }[q,n]g[m,n].
\end{align}
According to (\ref{eq:BasicFRFT16}), if $\Delta_u\Delta_x=\frac{2\pi|\sin\alpha|}{N}$, $\Delta_v\Delta_y=\frac{2\pi|\sin\beta|}{N}$, $\sin\alpha>0$ and $\sin\beta>0$, the 2D DFRFT can be
implemented by two 2D discrete CMs and one 2D DFT:
\begin{align}\label{eq:BasicFRFT20}
{\cal F}^{\:\alpha,\beta}_{x,y}=\frac{j}{{\sqrt { - \sin \alpha \sin \beta } }}{\cal O}_{\textmd{CM}}^{\bb H}\:
{\cal F}_{x,y}\:
{\cal O}_{\textmd{CM}}^{\bb H},
\end{align}
where
chirp matrix $\bb H$ is given by
\begin{align}\label{eq:BasicFRFT24}
\bb H=\bb A\bb B^{-1}=\bb B^{-1}\bb A= \begin{bmatrix}
{\cot \alpha }&0\\
0&{\cot \beta }
\end{bmatrix}.
\end{align}
One 2D FFT and two pointwise products are used in ${\cal F}^{\:\alpha,\beta}_{x,y}$ and thus  totally involve  $\frac{{{N^2}}}{2}{{\log }_2}{N^2}+2N^2$ complex multiplications.

\subsection{2D Discrete Affine Transformation}\label{subsec:BasicAffine}
When $\bb B$ and $\bb C$ are both $\bb 0$, one has $\bb A=(\bb D^T)^{-1}$ since ${\bf{A}}{{\bf{D}}^T}=\bb I$:
\begin{align}\label{eq:BasicAffine02}
\begin{bmatrix}
{\bf{A}}&{\bf{B}}\\
{\bf{C}}&{\bf{D}}
\end{bmatrix} = \begin{bmatrix}
(\bb D^T)^{-1}&{\bf{0}}\\
{\bf{0}}&{\bf{D}}
\end{bmatrix},
\quad\textmd{where}\quad
\bb D=\begin{bmatrix}
d_{11}&d_{12}\\
d_{21}&d_{22}
\end{bmatrix}.
\end{align}
From 
(\ref{eq:Intro20}), 
the above ABCD matrix leads to
\begin{align}\label{eq:BasicAffine04}
G(u,v)&= {\cal O}_{\textmd{NsLCT}}^{((\bb D^T)^{-1}, \bb 0;\bb 0,\bb D)}\{g(x,y)\}\nn\\
&=\sqrt {\det ({\bf{D}})} g({d_{11}}u + {d_{21}}v,{d_{12}}u + {d_{22}}v),
\end{align}
which is a 2D affine transformation. 
Sampling (\ref{eq:BasicAffine04}) with $\Delta_u$ and $\Delta_v$ yields
\begin{align}\label{eq:BasicAffine08}
G(p\Delta_u,q\Delta_v)=\sqrt {\det ({\bf{D}})} g({d_{11}}p\Delta_u + {d_{21}}q\Delta_v,{d_{12}}p\Delta_u + {d_{22}}q\Delta_v).
\end{align}
However, $G(p\Delta_u,q\Delta_v)$ 
is often not available when there are a limited number of input samples.
Accordingly, 2D interpolation is necessary.

Here, we introduce the bilinear interpolation method that is 
used in  Ding's method \cite{ding2012improved}.
With the discrete input $g[m,n]=g(m\Delta_x,n\Delta_y)$,
the 2D discrete affine transformation, denoted by ${\cal O}_{\textmd{Affine}}^{\bb D}$,
based on bilinear interpolation is given by
\begin{align}\label{eq:BasicAffine12}
&G[p,q]= {\cal O}_{\textmd{Affine}}^{\bb D}\{g[m,n]\}\nn\\
& \buildrel \Delta \over =\sqrt {\det ({\bf{D}})} \left\{ {(1 - k)(1 - l)\cdot g[{p_2},{q_2}] + (1 - k)l\cdot g[{p_2},{q_2} + 1]} \right.\nn\\
&\qquad\quad\ \ \left. { + k(1 - l) \cdot g[{p_2} + 1,{q_2}] + kl \cdot g[{p_2} + 1,{q_2} + 1]} \right\},
\end{align}
where
\begin{align}\label{eq:BasicAffine16}
&p_1=\frac{{d_{11}}p\Delta_u + {d_{21}}q\Delta_v}{\Delta_x},\quad q_1=\frac{{d_{12}}p\Delta_u + {d_{22}}q\Delta_v}{\Delta_y},\nn\\
&p_2=\left\lfloor {p_1} \right\rfloor, \ \ q_2=\left\lfloor {q_1} \right\rfloor,\ \ k=p_1-p_2, \ \ \textmd{and}\ \ l=q_1-q_2.
\end{align}
The symbol $\left\lfloor {\ } \right\rfloor$ denotes floor function.
In (\ref{eq:BasicAffine12}), if $(1-k)(1-l)$, $(1-k)l$, $k(1-l)$ and $kl$ are 
pre-computed, each output sample
$G[p,q]$ 
requires $8$ real multiplications, or equivalently $2$ complex multiplications.
Therefore, the total number of required complex multiplications is $2N^2$.

Smaller $\Delta_x$ and $\Delta_y$ can decrease the approximation error between $G[p,q]$ and $G(p\Delta_u,q\Delta_v)$, but the additional upsampling 
preprocess requires more computation time.

\section{Review of Previous Works}\label{sec:Rew}
Most digital implementation methods of the 1D FRFT can be extended to the 1D LCT. However, the 2D NsLCT is much more complicated. Most digital implementation methods of the 2D FRFT is not suitable for the 2D NsLCT or need extensive modification.
In this section, the implementation algorithms of Ko{\c{c}}'s 2D NsDLCT \cite{kocc2010fast} and  Ding's 2D NsDLCT
\cite{ding2012improved} are introduced .
These two methods are based on the additivity property of 2D NsLCT and decompositions of ABCD matrix.
Again, assume the discrete input and output are both of size $N\times N$.

\subsection{Ko{\c{c}}'s 2D NsDLCT \cite{kocc2010fast}}\label{subsec:Koc}
In \cite{kocc2010fast}, Ko{\c{c}}, Ozaktas and Hesselink proposed a fast algorithm to compute the 2D NsLCT.
It is based on the Iwasawa decomposition \cite{wolf2004geometric,alieva2005alternative} that decomposes the ABCD matrix into
\begin{align}\label{eq:Koc04}
\begin{bmatrix}
{\bf{A}}&{\bf{B}}\\
{\bf{C}}&{\bf{D}}
\end{bmatrix} = \begin{bmatrix}
{\bf{I}}&{\bf{0}}\\
{\bf{G}}&{\bf{I}}
\end{bmatrix}\begin{bmatrix}
{\bf{S}}&{\bf{0}}\\
{\bf{0}}&{{{\bf{S}}^{ - 1}}}
\end{bmatrix}\begin{bmatrix}
{\bf{X}}&{\bf{Y}}\\
{ - {\bf{Y}}}&{\bf{X}}
\end{bmatrix},
\end{align}
where
\begin{align}\label{eq:Koc08}
{\bf{S}} &= {({\bf{A}}{{\bf{A}}^T} + {\bf{B}}{{\bf{B}}^T})^{1/2}}, \ \ &{\bf{G}} &= ({\bf{C}}{{\bf{A}}^T} + {\bf{D}}{{\bf{B}}^T}){{\bf{S}}^{ - 2}},\nn\\
{\bf{X}} &= {{\bf{S}}^{ - 1}}{\bf{A}}\qquad\qquad \textmd{and} &{\bf{Y}} &= {{\bf{S}}^{ - 1}}{\bf{B}}.
\end{align}
In (\ref{eq:Koc04}), the first and second matrices correspond to 2D CM and 2D affine transformation, respectively.
The last matrix can be
further decomposed as follows:
\begin{align}\label{eq:Koc12}
\begin{bmatrix}
{\bf{X}}&{\bf{Y}}\\
{ - {\bf{Y}}}&{\bf{X}}
\end{bmatrix} = \begin{bmatrix}
{{{\bf{R}}_\phi}}&{\bf{0}}\\
{\bf{0}}&{{{\bf{R}}_\phi}}
\end{bmatrix}\begin{bmatrix}
{\bf{E}}&{\bf{F}}\\
{ - {\bf{F}}}&{\bf{E}}
\end{bmatrix}\begin{bmatrix}
{{{\bf{R}}_\theta}}&{\bf{0}}\\
{\bf{0}}&{{{\bf{R}}_\theta}}
\end{bmatrix},
\end{align}
where the first and third matrices correspond to 2D affine transformations (more precisely, geometric rotations) while the second matrix corresponds to 2D FRFT:
\begin{align}\label{eq:Koc16}
{{\bf{R}}_\theta } &= \begin{bmatrix}
{\cos \theta }&{\sin \theta }\\
{ - \sin \theta }&{\cos \theta }
\end{bmatrix},\quad
&{{\bf{R}}_\phi} &= \begin{bmatrix}
{\cos \phi }&{\sin \phi }\\
{ - \sin \phi }&{\cos \phi }
\end{bmatrix},\nn\\
{\bf{E}} &= \begin{bmatrix}
{\cos \alpha }&0\\
0&{\cos \beta }
\end{bmatrix},\quad
&{\bf{F}} &= \begin{bmatrix}
{\sin \alpha }&0\\
0&{\sin \beta }
\end{bmatrix}.
\end{align}
The fractional angles $\alpha$ and $\beta$ can be obtained from
\begin{align}\label{eq:Koc20}
\exp \left[ {j(\alpha  + \beta )} \right] &= \det \left( {{\bf{X}} + j{\bf{Y}}} \right),\nn\\
\cos (\alpha  - \beta ) &= \det \bb X + \det \bb Y.
\end{align}
And then the values of rotation angles $\theta$ and $\phi$ can be determined by solving the following two equations:
\begin{align}\label{eq:Koc24}
&\exp \left[ {j\left( {\theta  + \phi  + (\alpha  + \beta )/2} \right)} \right] \nn\\
&\ = \frac{{{X_{11}} + {X_{22}} - {Y_{12}} + {Y_{21}} + j\left( {{X_{12}} - {X_{21}} + {Y_{11}} + {Y_{22}}} \right)}}{{2\cos \left[ {(\alpha  - \beta )/2} \right]}}\nn\\
&\exp \left[ {j\left( {\theta  - \phi  + (\alpha  + \beta )/2} \right)} \right]\nn\\
&\ = \frac{{j\left( { - {X_{11}} + {X_{22}} + {Y_{12}} + {Y_{21}}} \right) + {X_{12}} + {X_{21}} + {Y_{11}} - {Y_{22}}}}{{2\sin \left[ {(\alpha  - \beta )/2} \right]}}.
\end{align}
Substituting  (\ref{eq:Koc12}) into (\ref{eq:Koc04}) leads to
\begin{align}\label{eq:Koc28}
\begin{bmatrix}
{\bf{A}}&{\bf{B}}\\
{\bf{C}}&{\bf{D}}
\end{bmatrix} = \underbrace{\begin{bmatrix}
{\bf{I}}&{\bf{0}}\\
{\bf{G}}&{\bf{I}}
\end{bmatrix}}_{\textmd{2D CM}}
\underbrace{\begin{bmatrix}
{{\bf{S}}{{\bf{R}}_\phi }}&{\bf{0}}\\
{\bf{0}}&{{{\bf{S}}^{ - 1}}{{\bf{R}}_\phi }}
\end{bmatrix}}_{\textmd{2D affine}}
\underbrace{\begin{bmatrix}
{\bf{E}}&{\bf{F}}\\
{ - {\bf{F}}}&{\bf{E}}
\end{bmatrix}}_{\textmd{2D FRFT}}
\underbrace{\begin{bmatrix}
{{{\bf{R}}_\theta }}&{\bf{0}}\\
{\bf{0}}&{{{\bf{R}}_\theta }}
\end{bmatrix}}_{\textmd{2D affine}}.
\end{align}
Based on this decomposition, Ko{\c{c}} \emph{et al.}'s developed a 2D NsDLCT, denoted by
${\cal O}_{\textmd{Ko{\c{c}}}}^{(\bb A, \bb B;\bb C,\bb D)}$, consisting of one 2D discrete CMs, two 2D discrete affine transformation and one 2D DFRFT:
\begin{align}\label{eq:Koc36}
{\cal O}_{\textmd{Ko{\c{c}}}}^{(\bb A, \bb B;\bb C,\bb D)}& \buildrel \Delta \over =
 {\cal O}_{\textmd{CM}}^{\bb G}\:
 {\cal O}_{\textmd{Affine}}^{{{{\bf{S}}^{ - 1}}{{\bf{R}}_\phi }}}\:
 {\cal F}^{\:\alpha,\beta}_{x,y}\:
 {\cal O}_{\textmd{Affine}}^{{{{\bf{R}}_\theta }}}\nn\\
 &= {\cal O}_{\textmd{CM}}^{\bb G}\:
 {\cal O}_{\textmd{Affine}}^{{{{\bf{S}}^{ - 1}}{{\bf{R}}_\phi }}}\:
{\cal O}_{\textmd{CM}}^{\bb H}\:
{\cal F}_{x,y}\:
{\cal O}_{\textmd{CM}}^{\bb H}\:
 {\cal O}_{\textmd{Affine}}^{{{{\bf{R}}_\theta }}}.
\end{align}
These basic discrete operations used above have been defined in (\ref{eq:BasicCM12}), (\ref{eq:BasicDFT16}), (\ref{eq:BasicDFT20}), (\ref{eq:BasicFRFT20}) and (\ref{eq:BasicAffine12}),
and $\bb H$ is given by
\begin{align}\label{eq:Koc40}
\bb H=\bb E\bb F^{-1}=\bb F^{-1}\bb E=\begin{bmatrix}
{\cot \alpha }&0\\
0&{\cot \beta }
\end{bmatrix}.
\end{align}
If the input of the 2D DFT ${\cal F}_{x,y}$ is zero-padded to $N'\times N'$ where $N'>N$, the output of the 2D DFT will have smaller sampling intervals.
Then, as mentioned in  Sec.~\ref{sec:Basic}.\ref{subsec:BasicAffine}, the interpolation in the 2D discrete affine transformation ${\cal O}_{\textmd{Affine}}^{{{{\bf{S}}^{ - 1}}{{\bf{R}}_\phi }}}$ will have higher accuracy.
But
the cost is higher computational complexity.
If two-times upsampling
($N'=2N$) is employed, 
the number of complex multiplications used in Ko{\c{c}}'s method becomes
\begin{align}\label{eq:Koc42}
&N^2+2N^2+N'^2+\frac{{{N'^2}}}{2}{{\log }_2}{N'^2}+N^2+2N^2\\
&=2N^2{\log_2}{N^2}+14N^2.\label{eq:Koc44}
\end{align}
The six terms in (\ref{eq:Koc42})
(from left to right) are the numbers of complex multiplications used in the six discrete operations
in (\ref{eq:Koc36})
(from left to right), respectively.

\subsection{Ding's 2D NsDLCT \cite{ding2012improved}}\label{subsec:Ding}
Since 2D discrete affine transformations will introduce interpolation error,
Ding, Pei and Liu proposed a more accurate 2D NsDLCT in which only one 2D discrete affine transformation is used.
In their work, the ABCD matrix is decomposed into another form:
\begin{align}\label{eq:Ding04}
\begin{bmatrix}
{\bf{A}}&{\bf{B}}\\
{\bf{C}}&{\bf{D}}
\end{bmatrix} \!=\!
\underbrace{
\begin{bmatrix}
{\bf{I}}&{\bf{0}}\\
{{\bf{D}}{{\bf{B}}^{ - 1}}}&{\bf{I}}
\end{bmatrix}}_{\textmd{2D CM}}\!
\underbrace{
\begin{bmatrix}
{\bf{B}}&{\bf{0}}\\
{\bf{0}}&{{{({{\bf{B}}^T})}^{ - 1}}}
\end{bmatrix}}_{\textmd{2D affine}}\!
\underbrace{
\begin{bmatrix}
{\bf{0}}&{\bf{I}}\\
{ - {\bf{I}}}&{\bf{0}}
\end{bmatrix}}_{\textmd{2D FT}}\!
\underbrace{
\begin{bmatrix}
{\bf{I}}&{\bf{0}}\\
{{{\bf{B}}^{ - 1}}{\bf{A}}}&{\bf{I}}
\end{bmatrix}}_{\textmd{2D CM}}.
\end{align}
Based on (\ref{eq:Ding04}), Ding \emph{et al.}'s developed another
type of 2D NsDLCT, denoted by ${\cal O}_{\textmd{Ding}}^{(\bb A, \bb B;\bb C,\bb D)}$,
which connects a 2D discrete CM, a 2D discrete affine transformation, a 2D DFT and another 2D discrete CM in series:
\begin{align}\label{eq:Ding12}
{\cal O}_{\textmd{Ding}}^{(\bb A, \bb B;\bb C,\bb D)}&=
 {\cal O}_{\textmd{CM}}^{{\bf{D}}{{\bf{B}}^{ - 1}}}\:
 {\cal O}_{\textmd{Affine}}^{{{({{\bf{B}}^T})}^{ - 1}}}\:
 {\cal F}_{x,y}\:
 {\cal O}_{\textmd{CM}}^{{{\bf{B}}^{ - 1}}{\bf{A}}}.
\end{align}
These basic discrete operations are  defined in (\ref{eq:BasicCM12}), (\ref{eq:BasicDFT16}) and (\ref{eq:BasicAffine12}).

Again, suppose  two-times upsampling is performed in 
$ {\cal F}_{x,y}$ for higher accuracy in the 2D discrete affine transformation ${\cal O}_{\textmd{Affine}}^{{{({{\bf{B}}^T})}^{ - 1}}}$.
Assume 
two-times upsampling ($N'=2N$) is employed.
The number of complex multiplications required by Ding's method is given by
\begin{align}\label{eq:Ding16}
{N^2} + 2{N^2} + \frac{{{{N'}^2}}}{2}{\log _2}{{N'}^2} + {N^2} = 2{N^2}{\log _2}{N^2}+8{N^2}.
\end{align}
Comparing (\ref{eq:Koc44}) and (\ref{eq:Ding16}),
Ding's method has lower computational complexity than  Ko{\c{c}}'s method.

\section{Proposed 2D CM-CC-CM-CC NsDLCTs}\label{sec:Proposed}
Since the 2D discrete affine transformation will introduce interpolation error, in this paper, some 2D NsDLCTs are developed with no 2D discrete affine transformation involved.
In \cite{koc2008digital}, Ko{\c{c}} \emph{et al.} have introduced a variety  of decompositions for 1D LCT, where the parameter matrix is decomposed into three, four or five matrices.
The CM-CC-CM decomposition \cite{koc2008digital,pei2011discrete} is the only one without using scaling operation.
Therefore, to avoid affine transformations in two dimensions, a possible method is decomposing the ABCD matrix 
into CM matrices $(\bb I, \bb 0;\bb C, \bb I)$ and CC matrices $(\bb I, \bb B;\bb 0, \bb I)$.
Accordingly, all the proposed methods
are composed of only the 2D discrete CMs and 2D discrete CCs. 
More precisely, only the 2D discrete CMs, 2D DFTs and 2D IDFTs are used.
These basic discrete operations have been defined in (\ref{eq:BasicCM12}), (\ref{eq:BasicDFT16}), (\ref{eq:BasicDFT20}) and (\ref{eq:BasicCC20}).

\subsection{2D NsDLCT Based on CM-CC-CM Decomposition When $\bb B=\bb B^T$ and $\det \bb B\neq0$}\label{subsec:Bsym}
First, suppose the 2D NsLCT can also be decomposed into CM-CC-CM.
Then, it implies that the ABCD matrix can be expressed as the following form:
\begin{align}\label{eq:Bsym04}
\begin{bmatrix}
{\bf{A}}&{\bf{B}}\\
{\bf{C}}&{\bf{D}}
\end{bmatrix} =
\underbrace{\begin{bmatrix}
{\bf{I}}&{\bf{0}}\\
({\bf{D}-\bb I){{\bf{B}}^{ - 1}}}&{\bf{I}}
\end{bmatrix}}_{\textmd{2D CM}}
\underbrace{\begin{bmatrix}
{\bf{I}}&{\bf{B}}\\
{\bf{0}}&{\bf{I}}
\end{bmatrix}}_{\textmd{2D CC}}
\underbrace{\begin{bmatrix}
{\bf{I}}&{\bf{0}}\\
{{{\bf{B}}^{ - 1}}(\bf{A}-\bb I})&{\bf{I}}
\end{bmatrix}}_{\textmd{2D CM}}.
\end{align}
Obviously, $\bb B$ must be invertible.
And since all the matrix in (\ref{eq:Bsym04}) should satisfy the constraints in (\ref{eq:Intro12}) or (\ref{eq:Intro16}), 
the necessary and sufficient condition is that $\bb B$ is symmetric.
Therefore, if
\begin{align}\label{eq:Bsym08}
\bb B=\bb B^T\quad \textmd{and}\quad \det \bb B\neq0,
\end{align}
the 2D NsDLCTs based on the CM-CC-CM decomposition is given by
\begin{align}\label{eq:Bsym12}
{\cal O}_{\textmd{NsDLCT}}^{(\bb A, \bb B;\bb C,\bb D)}&=
 {\cal O}_{\textmd{CM}}^{({\bf{D}-\bb I){{\bf{B}}^{ - 1}}}}
{\cal O}_{\textmd{CC}}^{\bb B}\:
 {\cal O}_{\textmd{CM}}^{{{{\bf{B}}^{ - 1}}(\bf{A}-\bb I})}\nn\\
 &= {\cal O}_{\textmd{CM}}^{({\bf{D}-\bb I){{\bf{B}}^{ - 1}}}}
{\cal F}_{x,y}^{-1}\:
{\cal O}_{\textmd{CM}}^{-\bb B}\:
{\cal F}_{x,y}\:
 {\cal O}_{\textmd{CM}}^{{{{\bf{B}}^{ - 1}}(\bf{A}-\bb I})}.
\end{align}
In this method, two 2D FFTs and three pointwise products are utilized 
and totally require $N^2\log_2N^2+3N^2$ complex multiplications. 
The 2D FRFT and gyrator transform \cite{rodrigo2007gyrator,pei2009properties} are the special cases of the 2D NsLCT with  $\bb B=\bb B^T$ and $ \det \bb B\neq0$, and thus can be digitally implemented by (\ref{eq:Bsym12}).

\subsection{2D NsDLCT Based on CM-CC-CM-CC Decomposition When $\bb B\neq\bb B^T$ or $\det \bb B=0$}\label{subsec:Bnonsym}
The CM-CC-CM decomposition in (\ref{eq:Bsym04}) is valid only when $\bb B\neq\bb B^T$ and $\det \bb B=0$.
Consider the more general case that ABCD matrix is arbitrary (but the constraints in (\ref{eq:Intro12}) or (\ref{eq:Intro16}) should be satisfied) and at most has ten degrees of freedom.
The CM matrix $(\bb I, \bb 0;\bb C, \bb I)$ and CC matrix $(\bb I, \bb B;\bb 0, \bb I)$ have only three degrees of freedom.
Four CM/CC matrices are required to 
describe the ABCD matrix.
Accordingly, the CM-CC-CM-CC decomposition that has twelve degrees of freedom is considered.
First, decompose the ABCD matrix into
\begin{align}\label{eq:Bnonsym04}
\begin{bmatrix}
{\bf{A}}&{\bf{B}}\\
{\bf{C}}&{\bf{D}}
\end{bmatrix} = \begin{bmatrix}
{\bf{A}}&{\bf{B}}'\\
{\bf{C}}&{\bf{D}}'
\end{bmatrix}\begin{bmatrix}
{\bf{I}}&{\bf{H}}\\
{\bf{0}}&{\bf{I}}
\end{bmatrix}
=\begin{bmatrix}
{\bf{A}}&\bb A\bb H+{\bf{B}}'\\
{\bf{C}}&\bb C\bb H+{\bf{D}}'
\end{bmatrix},
\end{align}
where $\bb H=\bb H^T$, $\bb B'=\bb B-\bb A\bb H$ and $\bb D'=\bb D-\bb C\bb H$.
If ${\bf{B}}'$ is symmetric and invertible, according to (\ref{eq:Bsym04}), the ABCD matrix can be further decomposed as follows:
\begin{align}\label{eq:Bnonsym08}
\begin{bmatrix}
{\bf{A}}&{\bf{B}}\\
{\bf{C}}&{\bf{D}}
\end{bmatrix} =
\underbrace{\begin{bmatrix}
{\bf{I}}&{\bf{0}}\\
({\bf{D}'-\bb I){{\bf{B}}'^{ - 1}}}&{\bf{I}}
\end{bmatrix}}_{\textmd{2D CM}}\!\!
\underbrace{\begin{bmatrix}
{\bf{I}}&{\bf{B}}'\\
{\bf{0}}&{\bf{I}}
\end{bmatrix}}_{\textmd{2D CC}}\!\!
\underbrace{\begin{bmatrix}
{\bf{I}}&{\bf{0}}\\
{{{\bf{B}}'^{ - 1}}(\bf{A}-\bb I})&{\bf{I}}
\end{bmatrix}}_{\textmd{2D CM}}\!\!
\underbrace{\begin{bmatrix}
{\bf{I}}&{\bf{H}}\\
{\bf{0}}&{\bf{I}}
\end{bmatrix}}_{\textmd{2D CC}}.
\end{align}
This decomposition is valid even if $\bb B\neq\bb B^T$ or $\det \bb B=0$.
The 2D NsDLCT based on the above CM-CC-CM-CC decomposition is given by
\begin{align}\label{eq:Bnonsym12}
{\cal O}_{\textmd{NsDLCT}}^{(\bb A, \bb B;\bb C,\bb D)}=
 {\cal O}_{\textmd{CM}}^{({\bf{D}'-\bb I){{\bf{B}'}^{ - 1}}}}
{\cal O}_{\textmd{CC}}^{\bb B'}\:
 {\cal O}_{\textmd{CM}}^{{{{\bf{B}}'^{ - 1}}(\bf{A}-\bb I})}\:
 {\cal O}_{\textmd{CC}}^{\bb H}.
\end{align}
Since the CM-CC-CM-CC decomposition has two more degrees of freedom, there are infinite number of 
possible decomposition results
(i.e. infinite choices of $\bb B'$,  $\bb D'$  and $\bb H$).

There are two approaches to determine $\bb B'$,  $\bb D'$  and $\bb H$.
Firstly, if
$\bb A$ is invertible, once $\bb B'$ is determined, $\bb H$ can be obtained from $\bb H=\bb A^{-1}(\bb B-\bb B')$ and then $\bb D'$ from $\bb D'=\bb D-\bb C\bb H$.
A valid $\bb B'$ for the CM-CC-CM-CC decomposition should satisfy the following three conditions:
\begin{align}\label{eq:Bnonsym16}
\bb B'=\bb B'^T,\quad \det\bb B'\neq0\quad \textmd{and} \quad{\bf{A}}{{\bf{B}}'^T} = {\bf{B}}'{{\bf{A}}^T};
\end{align}
that is,
\begin{align}\label{eq:Bnonsym20}
&\bb B'=\begin{bmatrix}
b'_{11}&b'_{12}\\
b'_{12}&b'_{22}
\end{bmatrix},\qquad b'_{11}b'_{22}-b'^2_{12}\neq0,\\
&\textmd{and}\qquad a_{11}b'_{12}+a_{12}b'_{22}=a_{21}b'_{11}+a_{22}b'_{12}.\label{eq:Bnonsym24}
\end{align}
Since $(\bb A, \bb B;\bb C,\bb D)$ and $(\bb A, \bb B';\bb C,\bb D')$ need to satisfy the constraints in (\ref{eq:Intro12}), it is required that
${\bf{A}}{{\bf{B}}^T} = {\bf{B}}{{\bf{A}}^T}$ and ${\bf{A}}{{\bf{B}}'^T} = {\bf{B}}'{{\bf{A}}^T}$.
Thus, one has ${\bf{A}}(\bb B-{\bf{B}}')^T = (\bb B-\bf{B}'){{\bf{A}}^T}$, and it follows that $\bb H=\bb H^T$:
\begin{align}\label{eq:Bnonsym28}
\bb H= \bb A^{-1}(\bb B-\bf{B}')=(\bb B-{\bf{B}}')^T(\bb A^{-1})^T =\bb H^T.
\end{align}
Secondly,
if $\bb A$ is non-invertible, one has to determine $\bb H$ 
first, and then obtain $\bb B'$ and $\bb D'$ from $\bb B'=\bb B-\bb A\bb H$ and $\bb D'=\bb D-\bb C\bb H$, respectively.
A valid $\bb H$ should lead to
\begin{align}\label{eq:Bnonsym32}
\bb H=\bb H^T,\quad \bb B'=\bb B'^T\quad \textmd{and} \quad \det\bb B'\neq0.
\end{align}
${\bf{A}}{{\bf{B}}'^T} = {\bf{B}}'{{\bf{A}}^T}$ doesn't need to be considered   because it is true when $\bb H=\bb H^T$ is true:
\begin{align}\label{eq:Bnonsym36}
{\bf{A}}{{\bf{B}}'^T}={\bf{A}}{{\bf{B}}^T}-\bb A\bb H^T\bb A^T = {\bf{B}}{{\bf{A}}^T}-\bb A\bb H\bb A^T = {\bf{B}}'{{\bf{A}}^T}.
\end{align}

Since there are infinite number of solutions of $\bb H$ to (\ref{eq:Bnonsym32}) (or $\bb B'$ to (\ref{eq:Bnonsym16})),
the problem is what the best choice of $\bb H$  (or $\bb B'$) is.
In the following, two types of 2D NsDLCT are proposed based on two types of $\bb H$.
One is for high accuracy while the other one is for low complexity.


\subsubsection{2D High-Accuracy NsDLCT (HA-NsDLCT)}\label{subsubsec:HA}

It has been shown in (\ref{eq:Intro04}) that the 2D NsLCT will produce an affine transformation in the space-spatial-frequency plane.
The 2D CM
with chirp matrix
$\bb C=\bb C^T=(c_{11},c_{12};c_{12},c_{22})$
will produce shearing in spatial-frequency domain:
\begin{align}\label{eq:HA12}
{\omega _u} &= ({c_{11}}x + {c_{12}}y) + {\omega _x},\nn\\
{\omega _v} &= ({c_{12}}x + {c_{22}}y) + {\omega _y}.
\end{align}
Consider the simple case that the input signal occupies $-S/2<x,y<S/2$ in space domain and $-S/2<\omega_x,\omega_y<S/2$ in spatial-frequency domain, i.e. has
space-spatial-bandwidth product $S^4$.
After the shearing in (\ref{eq:HA12}), the
space-spatial-bandwidth product becomes $\gamma({\bb C})\cdot S^4$, where the ratio function $\gamma(\cdot)$ is defined as
\begin{align}\label{eq:HA16}
\gamma({\bb C})=\left(|c_{11}| + |c_{12}| + 1\right)\left(|c_{12}| + |c_{22}|+1\right).
\end{align}
From (\ref{eq:Intro04}), the 2D CC 
with chirp matrix $\bb B=\bb B^T=(b_{11},b_{12};b_{12},b_{22})$
will lead to shearing in space domain:
\begin{align}\label{eq:HA20}
u &= x + ({b_{11}}{\omega _x} + {b_{12}}{\omega _y}),\nn\\
v &= y + ({b_{12}}{\omega _x} + {b_{22}}{\omega _y}).
\end{align}
The
space-spatial-bandwidth product $S^4$ through the 2D CC becomes $\gamma({\bb B})\cdot S^4$.
Too large
space-spatial-bandwidth product will yield serious overlapping and aliasing effects.

In order to minimize the increase of space-spatial-bandwidth product caused by the two CMs and two CCs in (\ref{eq:Bnonsym12}), $\bb H$ is determined by
the following optimization problem:
\begin{align}\label{eq:HA24}
\mathop {\min }\limits_{\bb H}\ \ &\gamma\! \left( {({\bf{D'}} - {\bf{I}}){{\bf{B}}^\prime }^{ - {\bf{1}}}} \right) \cdot \gamma\! \left( {{{\bf{B}}^\prime }} \right) \cdot \gamma\! \left( {{{\bf{B}}^\prime }^{ - 1}({\bf{A}} - {\bf{I}})} \right) \cdot \gamma\! \left( {\bf{H}} \right)\\
\ \textmd{s.t.}\ \ \ &\quad \bb H=\bb H^T,\quad \bb B'=\bb B'^T\quad \textmd{and} \quad \det\bb B'\neq0,\nn
\end{align}
where $\bb B'=\bb B-\bb A\bb H$ and $\bb D'=\bb D-\bb C\bb H$.
The resulting 2D NsDLCT
is called 2D high-accuracy NsDLCT (HA-NsDLCT) and given by
\begin{align}\label{eq:HA28}
&{\cal O}_{\textmd{HA-NsDLCT}}^{(\bb A, \bb B;\bb C,\bb D)}=
 {\cal O}_{\textmd{CM}}^{({\bf{D}'-\bb I){{\bf{B}'}^{ - 1}}}}
{\cal O}_{\textmd{CC}}^{\bb B'}\:
 {\cal O}_{\textmd{CM}}^{{{{\bf{B}}'^{ - 1}}(\bf{A}-\bb I})}\:
 {\cal O}_{\textmd{CC}}^{\bb H}\nn\\
 &= {\cal O}_{\textmd{CM}}^{({\bf{D}'-\bb I){{\bf{B}'}^{ - 1}}}}
{\cal F}_{x,y}^{-1}\:
{\cal O}_{\textmd{CM}}^{-\bb B'}\:
{\cal F}_{x,y}\:
{\cal O}_{\textmd{CM}}^{{{{\bf{B}'}^{ - 1}}(\bf{A}-\bb I})}\:
{\cal F}_{x,y}^{-1}\:
{\cal O}_{\textmd{CM}}^{-\bb H}\:
{\cal F}_{x,y}.
\end{align}
Four 2D FFTs and four pointwise products are used and totally require 
\begin{align}\label{eq:HA32}
4\cdot \frac{1}{2}N^2\log_2N^2+4\cdot N^2=2N^2\log_2N^2+4N^2
\end{align}
complex multiplications.
The computational complexity of solving the optimization problem in (\ref{eq:HA24}) is negligible when $N$ is large enough.
From (\ref{eq:Koc44}), (\ref{eq:Ding16}) and (\ref{eq:HA32}), we can find out that
the 2D HA-NsDLCT has lower complexity than
Ko{\c{c}}'s method \cite{kocc2010fast} and Ding's method  \cite{ding2012improved}.

\subsubsection{2D Low-Complexity NsDLCT (LC-NsDLCT)}\label{subsubsec:LC}
To achieve lower computational complexity,
a symmetric $\bb H$ of either of the following two forms is considered:
\begin{align}\label{eq:LC04}
\bb H=\begin{bmatrix}
h_{11}&h_{12}\\
h_{12}&h_{22}
\end{bmatrix}=\begin{bmatrix}
h&0\\
0&0
\end{bmatrix}\quad \textmd{or}\quad
\begin{bmatrix}
0&0\\
0&h
\end{bmatrix}.
\end{align}
The above assumption is based on the fact that the CM-CC-CM-CC decomposition in (\ref{eq:Bnonsym08}) has two more degrees of freedom than the ABCD matrix.
Therefore, the number of variables in $\bb H$ can be reduced to one for lower complexity.
The 2D CC with $\bb H$ in (\ref{eq:LC04}) will reduce
into  1D CC with chirp rate $h$ operating in the $x$ direction or $y$ direction.
We call this type of 2D NsDLCT as 2D low-complexity NsDLCT (LC-NsDLCT).


A valid $\bb H$ should satisfy the constraints in (\ref{eq:Bnonsym32}).
The matrix $\bb B'=\bb B-\bb A\bb H$ needs to be symmetric.
Thus, $h$ in (\ref{eq:LC04})
is given by
\begin{align}\label{eq:LC08}
h=(b_{21}-b_{12})/a_{21}, \ \ \ & \textmd{when}\ \ \ \bb H=(h,0;0,0),\\
h=(b_{12}-b_{21})/a_{12}, \ \ \ & \textmd{when}\ \ \ \bb H=(0,0;0,h).\label{eq:LC10}
\end{align}
If both (\ref{eq:LC08}) and (\ref{eq:LC10}) are available, we can use the criterion in (\ref{eq:HA24}) to choose the one having 
higher accuracy.

There are two cases that the 2D LC-NsDLCT is not suitable.
When  $a_{12}=a_{21}=0$, both (\ref{eq:LC08}) and (\ref{eq:LC10}) are invalid.
In this case, another type of decomposition is used: $(\bb A, \bb B;\bb C, \bb D)=(-\bb B, \bb A;-\bb D, \bb C)(\bb 0, -\bb I;\bb I, \bb 0)$, where $(-\bb B, \bb A;-\bb D, \bb C)$ can be decomposed into CM-CC-CM when $\bb A$ is symmetric. The corresponding 2D NsDLCT has lower complexity than the 2D LC-NsDLCT.
Another case is
that neither of the two forms of $\bb H$ in (\ref{eq:LC04}) can yield invertible $\bb B'$.
For example, when $\bb B=\bb 0$, (\ref{eq:LC08}) and (\ref{eq:LC10}) yield $\bb H=\bb 0$, and it follows that $\bb B'=\bb B-\bb A\bb H=\bb 0$ is invertible.
Accordingly, if the assumption in (\ref{eq:LC04}) leads to $\det \bb B'=0$,
the 2D HA-NsDLCT in Sec.~\ref{sec:Proposed}.\ref{subsubsec:HA} is utilized instead.

Denote ${\cal F}_{x}$ and ${\cal F}_{x}^{-1}$ as 1D DFT and 1D IDFT applied in the  $x$ direction, respectively.
When $\bb H=(h,0;0,0)$,
the 2D LC-NsDLCT is given by
\begin{align}\label{eq:LC12}
&{\cal O}_{\textmd{LC-NsDLCT}}^{(\bb A, \bb B;\bb C,\bb D)}=
 {\cal O}_{\textmd{CM}}^{({\bf{D}'-\bb I){{\bf{B}'}^{ - 1}}}}
{\cal O}_{\textmd{CC}}^{\bb B'}\:
 {\cal O}_{\textmd{CM}}^{{{{\bf{B}}'^{ - 1}}(\bf{A}-\bb I})}\:
 {\cal O}_{\textmd{CC}}^{\bb H}\nn\\
 &= {\cal O}_{\textmd{CM}}^{({\bf{D}'-\bb I){{\bf{B}'}^{ - 1}}}}
{\cal F}_{x,y}^{-1}\:
{\cal O}_{\textmd{CM}}^{-\bb B'}\:
{\cal F}_{x,y}\:
{\cal O}_{\textmd{CM}}^{{{{\bf{B}'}^{ - 1}}(\bf{A}-\bb I})}\:
{\cal F}_{x}^{-1}\:
{\cal O}_{\textmd{CM}}^{-\bb H}\:
{\cal F}_{x},
\end{align}
where the last 2D discrete CM $ {\cal O}_{\textmd{CC}}^{\bb H}$ reduces into 1D discrete CM in the $x$ direction.
When $\bb H=(0,0;0,h)$, the ${\cal F}_{x}$ and ${\cal F}_{x}^{-1}$ in the above equation are  replaced by 
${\cal F}_{y}$ and ${\cal F}_{y}^{-1}$, respectively.

The 2D DFT/IDFT is equivalent to performing two 1D DFTs/IDFTs in  $x$ and $y$ directions.
It implies that ${\cal F}_{x}$ and ${\cal F}_{x}^{-1}$ (or ${\cal F}_{y}$ and ${\cal F}_{y}^{-1}$) have half the computational complexity of ${\cal F}_{x,y}$ and ${\cal F}_{x,y}^{-1}$, respectively.
Accordingly, the total number of complex multiplications used in the 2D LC-NsDLCT is
\begin{align}\label{eq:LC16}
&2\cdot \frac{1}{2}N^2\log_2N^2+4\cdot N^2+2\cdot \frac{1}{2}\cdot \frac{1}{2}N^2\log_2N^2\nn\\
&=\frac{3}{2}N^2\log_2N^2+4N^2.
\end{align}
Compared with (\ref{eq:HA32}), the 2D LC-NsDLCT has computational complexity $\frac{1}{2}N^2\log_2N^2$ lower than the 2D HA-NsDLCT.
The complexity of the 2D FFT is also $\frac{1}{2}N^2\log_2N^2$.
Since the 2D HA-NsDLCT involves four 2D FFTs and four pointwise products, we can say that the 2D LC-NsDLCT has complexity equivalent to three 2D FFTs and four pointwise products.
Although the 2D LC-NsDLCT features low complexity, later we will show that it also has higher accuracy than
Ko{\c{c}}'s method \cite{kocc2010fast} and Ding's method  \cite{ding2012improved}.

\begin{table*}[t]
\small
\begin{center}
\setstretch{1.5}
\caption{Complexity of Ko{\c{c}}'s method \cite{kocc2010fast},  Ding's method  \cite{ding2012improved}, proposed 2D HA-NsDLCT and proposed 2D LC-NsDLCT}\label{tab:table1}
\begin{tabular}{|c|c|c|c|c|}
\hline
 &No. of discrete affine transformations & No. of discrete CMs & No. of 2D FFTs & Total no. of complex mulplications\\
\hline\hline
Ko{\c{c}}'s method \cite{kocc2010fast} & 2 & 3 & 1& $2N^2{\log_2}{N^2}+14N^2$\\
\hline
Ding's method \cite{ding2012improved} & 1 & 2 & 1& $2{N^2}{\log _2}{N^2}+8{N^2}$\\
\hline
2D HA-NsDLCT & 0 & 4 & 4& $2N^2\log_2N^2+4N^2$\\
\hline
2D LC-NsDLCT  & 0 & 4 & 3\textsuperscript{$*$}& $\frac{3}{2}N^2\log_2N^2+4N^2$\\
\hline
\multicolumn{5}{l}{
\textsuperscript{$*$}\small{Two 2D FFTs and two 1D FFTs are used, and thus have complexity equivalent to three 2D FFTs. Refer to
the end of Sec.~\ref{sec:Proposed}.\ref{subsubsec:LC} for more details. }}
\end{tabular}
\end{center}
\vspace*{-10pt}
\end{table*}

\subsection{2D NsDLCT Based on CC-CM-CC-CM Decomposition When $\bb B\neq\bb B^T$ or $\det \bb B=0$}\label{subsec:CCCMCCCM}
If we want to perfectly reconstruct $g[m,n]$ from $G[p,q]$, the inverse transforms of the 2D HA-NsDLCT in (\ref{eq:HA28}) and 2D HA-NsDLCT in (\ref{eq:LC12}) should be of the following form:
\begin{align}\label{eq:CCCMCCCM04}
 {\cal O}_{\textmd{CC}}^{-\bb H}\:
  {\cal O}_{\textmd{CM}}^{-{\bb B'^{ - 1}}(\bb A-\bb I)}\:
  {\cal O}_{\textmd{CC}}^{-\bb B'}\:
 {\cal O}_{\textmd{CM}}^{-(\bb D'-\bb I){\bb B'^{ - 1}}}.
\end{align}
Accordingly, another type of decomposition, called CC-CM-CC-CM decomposition, is introduced.
First, decompose the ABCD matrix, say $(\bb A_1, \bb B_1;\bb C_1,\bb D_1)$,  into
\begin{align}\label{eq:CCCMCCCM08}
\begin{bmatrix}
{\bb {A}}_1&{\bb {B}}_1\\
{\bb {C}}_1&{\bb {D}}_1
\end{bmatrix} \!=\! \begin{bmatrix}
{\bb {I}}&\bb H_1\\
{\bb {0}}&{\bb {I}}
\end{bmatrix}\!\begin{bmatrix}
\bb A'_1&\bb B'_1\\
\bb C_1&\bb D_1
\end{bmatrix}
\!=\!\begin{bmatrix}
\bb A'_1+\bb H_1\bb C_1&\bb B'_1+\bb H_1\bb D_1\\
\bb C_1&\bb D_1
\end{bmatrix},
\end{align}
where $\bb H_1=\bb H_1^T$, $\bb B'_1=\bb B_1-\bb H_1\bb D_1$ and $\bb A'_1=\bb A_1-\bb H_1\bb C_1$.
If $\bb B'_1$ is symmetric and invertible, according to  (\ref{eq:Bsym04}), the ABCD matrix can be further decomposed as follows:
\begin{align}\label{eq:CCCMCCCM12}
\begin{bmatrix}
\bb A_1&\bb B_1\\
\bb C_1&\bb D_1
\end{bmatrix} &\!=\!
\underbrace{\begin{bmatrix}
{\bb {I}}&\bb H_1\\
{\bb {0}}&\bb I
\end{bmatrix}}_{\textmd{2D CC}}\!
\underbrace{\begin{bmatrix}
{\bb {I}}&{\bb {0}}\\
(\bb D_1-\bb I){\bb B_1'^{ - 1}}&{\bb {I}}
\end{bmatrix}}_{\textmd{2D CM}}\!
\underbrace{\begin{bmatrix}
{\bb {I}}&\bb B_1'\\
{\bb {0}}&{\bb {I}}
\end{bmatrix}}_{\textmd{2D CC}}\nn\\
&\qquad\qquad\qquad\qquad\qquad\times\!
\underbrace{\begin{bmatrix}
{\bb {I}}&{\bb {0}}\\
{\bb B_1'^{ - 1}}(\bb A'_1-\bb I)&{\bb {I}}
\end{bmatrix}}_{\textmd{2D CM}}.
\end{align}
Based on the CC-CM-CC-CM decomposition above, the 2D NsDLCT can be designed as
\begin{align}\label{eq:CCCMCCCM16}
{\cal O}_{\textmd{NsDLCT}}^{(\bb A_1, \bb B_1;\bb C_1,\bb D_1)}=
 {\cal O}_{\textmd{CC}}^{\bb H_1}\:
 {\cal O}_{\textmd{CM}}^{(\bb D_1-\bb I){\bb B_1'^{ - 1}}}\:
{\cal O}_{\textmd{CC}}^{\bb B'_1}\:
 {\cal O}_{\textmd{CM}}^{{\bb B_1'^{ - 1}}(\bb A_1'-\bb I)}.
\end{align}
Again, there are infinite number of possible decompositions.
So next we will prove that when $(\bb A_1, \bb B_1;\bb C_1,\bb D_1)=(\bb A, \bb B;\bb C,\bb D)^{-1}$, (\ref{eq:CCCMCCCM16}) will become (\ref{eq:CCCMCCCM04}) if $\bb H_1=-\bb H$.

Since the  ABCD matrix is given by
\begin{align}\label{eq:CCCMCCCM20}
\begin{bmatrix}
{\bb {A}}_1&{\bb {B}}_1\\
{\bb {C}}_1&{\bb {D}}_1
\end{bmatrix}=\begin{bmatrix}
{\bb {A}}&{\bb {B}}\\
{\bb {C}}&{\bb {D}}
\end{bmatrix}^{-1}=\begin{bmatrix}
{\bb {D}}^T&-{\bb {B}}^T\\
-{\bb {C}}^T&{\bb {A}}^T
\end{bmatrix},
\end{align}
one has
\begin{align}\label{eq:CCCMCCCM23}
\bb B'_1&=\bb B_1-\bb H_1\bb D_1=-\bb B^T-\bb H_1\bb A^T,\nn\\
\bb A'_1&=\bb A_1-\bb H_1\bb C_1=\bb D^T+\bb H_1\bb C^T.
\end{align}
Recall $\bb H=\bb H^T$, $\bb B'=\bb B-\bb A\bb H$ and $\bb D'=\bb D-\bb C\bb H$ used in the CM-CC-CM-CC decomposition in (\ref{eq:Bnonsym08}).
If $\bb H_1=-\bb H=-\bb H^T$, (\ref{eq:CCCMCCCM23}) becomes
\begin{align}\label{eq:CCCMCCCM24}
\bb B'_1&=-\bb B^T+\bb H^T\bb A^T=-\bb B'^T,\nn\\
\bb A'_1&=\bb D^T-\bb H^T\bb C^T=\bb D'^T.
\end{align}
The chirp matrices used in the four 2D CMs and CCs in (\ref{eq:CCCMCCCM16}) must be symmatric, and thus (\ref{eq:CCCMCCCM16}) can be written as
\begin{align}\label{eq:CCCMCCCM26}
{\cal O}_{\textmd{NsDLCT}}^{(\bb A_1, \bb B_1;\bb C_1,\bb D_1)}=
 {\cal O}_{\textmd{CC}}^{\bb H_1^T}\:
 {\cal O}_{\textmd{CM}}^{{(\bb B_1'^T)^{ - 1}}(\bb D_1^T-\bb I)}\:
{\cal O}_{\textmd{CC}}^{\bb B_1'^T}\:
 {\cal O}_{\textmd{CM}}^{(\bb A_1'^T-\bb I){(\bb B_1'^T)^{ - 1}}}.
\end{align}
From (\ref{eq:CCCMCCCM20}), (\ref{eq:CCCMCCCM24}) and $\bb H_1=-\bb H=-\bb H^T$, the above eqution can be rewritten as
\begin{align}\label{eq:CCCMCCCM30}
{\cal O}_{\textmd{NsDLCT}}^{(\bb A, \bb B;\bb C,\bb D)^{-1}}=
 {\cal O}_{\textmd{CC}}^{-\bb H}\:
 {\cal O}_{\textmd{CM}}^{-\bb B'^{-1}({\bb {A}}-\bb I)}\:
{\cal O}_{\textmd{CC}}^{-\bb B'}\:
 {\cal O}_{\textmd{CM}}^{-(\bb D'-\bb I)\bb B'^{-1}},
\end{align}
which is the same as (\ref{eq:CCCMCCCM04}) and thus can be used as the inverse transform of the 2D NsDLCT based on CM-CC-CM-CC decomposition.

Similarly, one can develop 2D HA-NsDLCT and 2D LC-NsDLCT based on the CC-CM-CC-CM decomposition.
With $\bb H_1$  determined by approaches similar to (\ref{eq:HA24}), (\ref{eq:LC08}) and (\ref{eq:LC10}), one has
\begin{align}\label{eq:CCCMCCCM40}
{\cal O}_{\textmd{HA-NsDLCT}}^{(\bb A_1, \bb B_1;\bb C_1,\bb D_1)}&={\cal F}_{x,y}^{-1}\:
 {\cal O}_{\textmd{CM}}^{-\bb H_1}\:
 {\cal F}_{x,y}\:
 {\cal O}_{\textmd{CM}}^{(\bb {D}_1-\bb I){{\bb {B}}_1'^{ - 1}}}\nn\\
&\qquad\qquad\qquad
 {\cal F}_{x,y}^{-1}\:
{\cal O}_{\textmd{CM}}^{-\bb B'_1}\:
{\cal F}_{x,y}\:
 {\cal O}_{\textmd{CM}}^{{{\bb {B}_1}'^{ - 1}}(\bb {A}_1'-\bb I)},\\
{\cal O}_{\textmd{LC-NsDLCT}}^{(\bb A_1, \bb B_1;\bb C_1,\bb D_1)}&={\cal F}_{x}^{-1}\:
 {\cal O}_{\textmd{CM}}^{-\bb H_1}\:
 {\cal F}_{x}\:
 {\cal O}_{\textmd{CM}}^{(\bb {D}_1-\bb I){\bb B_1'^{ - 1}}}\nn\\
&\qquad\qquad\qquad
 {\cal F}_{x,y}^{-1}\:
{\cal O}_{\textmd{CM}}^{-\bb B'_1}\:
{\cal F}_{x,y}\:
 {\cal O}_{\textmd{CM}}^{{{\bb {B}_1}'^{ - 1}}(\bb {A}_1'-\bb I)}.\label{eq:CCCMCCCM44}
\end{align}
Note that ${\cal F}_{x}$ and ${\cal F}_{x}^{-1}$ in the
2D LC-NsDLCT are  replaced by ${\cal F}_{y}$ and ${\cal F}_{y}^{-1}$, respectively, if $\bb H_1$ is of the form $(0,0;0,h_1)$.

\subsection{2D HA-NsDLCT and 2D LC-NsDLCT With Perfect Reversibility Property}\label{subsec:Reversible}
Like the continuous 2D NsLCT, the reversibility property for 2D NsDLCT is defined as
\begin{align}\label{eq:Reversible04}
{\cal O}_{\textmd{NsDLCT}}^{(\bb A, \bb B;\bb C,\bb D)^{-1}}{\cal O}_{\textmd{NsDLCT}}^{(\bb A, \bb B;\bb C,\bb D)}\{g[m,n]\}=g[m,n].
\end{align}
(\ref{eq:Bnonsym12}) and (\ref{eq:CCCMCCCM30}) show the 2D NsDLCT based on "CC-CM-CC-CM decomposition" can be used as the inverse transform of the 2D NsDLCT based on "CM-CC-CM-CC decomposition", and in fact vise versa.
In order to let ${\cal O}_{\textmd{NsDLCT}}^{(\bb A, \bb B;\bb C,\bb D)^{-1}}$ and ${\cal O}_{\textmd{NsDLCT}}^{(\bb A, \bb B;\bb C,\bb D)}$ use different
types of decompositions, we make the following assumption:
\begin{align}\label{eq:Reversible08}
&{\cal O}_{\textmd{NsDLCT}}^{(\bb A, \bb B;\bb C,\bb D)}\nn\\
&=\left\{ {\begin{array}{*{20}{l}}
{{\cal O}_{\textmd{CM}}^{({\bf{D}'-\bb I){{\bf{B}'}^{ - 1}}}}\:
{\cal O}_{\textmd{CC}}^{\bb B'}\:
 {\cal O}_{\textmd{CM}}^{{{{\bf{B}}'^{ - 1}}(\bf{A}-\bb I})}\:
 {\cal O}_{\textmd{CC}}^{\bb H},}&{\ \textmd{for}\  \ \textmd{tr}(\bb B)>0}\\
{ {\cal O}_{\textmd{CC}}^{\bb H}\:
 {\cal O}_{\textmd{CM}}^{(\bb D-\bb I){\bb B'^{ - 1}}}\:
{\cal O}_{\textmd{CC}}^{\bb B'}\:
 {\cal O}_{\textmd{CM}}^{{\bb B'^{ - 1}}(\bb A'-\bb I)},}&{\ \textmd{for}\ \ \textmd{tr}(\bb B)<0}
\end{array}} \right.,
\end{align}
where $\textmd{tr}(\cdot)$ denotes matrix trace.
For example, 
given some $(\bb A, \bb B;\bb C,\bb D)$ where $\textmd{tr}(\bb B)<0$, the forward transform
${\cal O}_{\textmd{NsDLCT}}^{(\bb A, \bb B;\bb C,\bb D)}$ will use the second type of (\ref{eq:Reversible08}).
Since $(\bb A, \bb B;\bb C,\bb D)^{-1}=(\bb D^T, -\bb B^T;-\bb C^T,\bb A^T)$ leads to $\textmd{tr}(-\bb B^T)>0$, the inverse transform  ${\cal O}_{\textmd{NsDLCT}}^{(\bb A, \bb B;\bb C,\bb D)^{-1}}$ will use the first type.
Then, perfect reconstruction is achieved.

Replacing (\ref{eq:Reversible08}) by (\ref{eq:HA28}) and (\ref{eq:CCCMCCCM40}) leads to
the reversible 2D HA-NsDLCT.
Replacing (\ref{eq:Reversible08}) by (\ref{eq:LC12}) and (\ref{eq:CCCMCCCM44}) leads to
the reversible 2D LC-NsDLCT.

\section{Comparisons Between Proposed 2D NsDLCTs and Previous Works}\label{sec:Comp}
In this section, the proposed 2D HA-NsDLCT and 2D LC-NsDLCT will be compared with Ko{\c{c}}'s method \cite{kocc2010fast} and  Ding's 2D method \cite{ding2012improved} in computational complexity, accuracy, additivity property and reversibility property.
The computational complexity has been analyzed in (\ref{eq:Koc44}), (\ref{eq:Ding16}), (\ref{eq:HA32}) and (\ref{eq:LC16}), and is summarized in TABLE~\ref{tab:table1}.

\begin{figure}[t]
\centering
\includegraphics[width=1\columnwidth,clip=true]{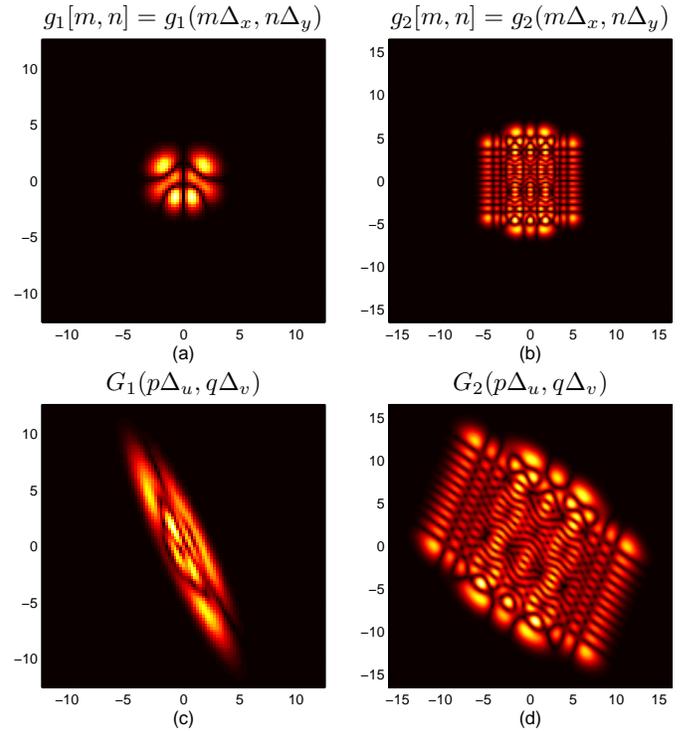}
\vspace*{-12pt}
\caption{
The 2D discrete input signals: (a) $100\times100$ $g_1[m,n]$ with $\Delta_x=\Delta_y=0.25$ (sampled $HG_{1,2}+HG_{3,1}$), and (b) $165\times165$ $g_2[m,n]$ with $\Delta_x=\Delta_y=0.2$ (sampled $HG_{2,18}+HG_{14,11}$).
The approximate sampled 2D NsLCTs calculated by direct method in (\ref{eq:Intro32}): (c) $100\times100$ approximate $G_1(p\Delta_u,q\Delta_v)$ with $\Delta_u=\Delta_v=0.25$ and $(\bb A_1, \bb B_1;\bb C_1,\bb D_1)$ given in (\ref{eq:Acc24}), and (d)
$165\times165$ approximate $G_2(p\Delta_u,q\Delta_v)$ with $\Delta_u=\Delta_v=0.2$ and $(\bb A_2, \bb B_2;\bb C_2,\bb D_2)$ given in (\ref{eq:Acc28}).
The input is not zero-padded in these two examples.
}
\label{fig:InputHG}
\vspace*{-4pt}
\end{figure}

\subsection{Accuracy}\label{subsec:Acc}
Assume $G(u,v)$ is the 2D NsLCT of  input signal $g(x,y)$. 
Given $g[m,n] \buildrel \Delta \over = g(m\Delta_x,n\Delta_y)$ as the discrete input,
a highly accurate 2D NsDLCT should have discrete output approximating the sampled 2D NsLCT, i.e. $G(p\Delta_u,q\Delta_v)$, with very small error.
Accordingly, the accuracy is measured by the normalized mean-square error (NMSE) 
defined as
\begin{align}\label{eq:Acc04}
{\textmd{NMSE}}_{\textmd{acc}} = \frac{\sum\limits_p^{}\sum\limits_q^{} {\left|G[p,q]-G(p\Delta_u,q\Delta_v) \right|}^2 }{\sum\limits_p^{}\sum\limits_q^{} {\left|G(p\Delta_u,q\Delta_v) \right|}^2 },
\end{align}
where $G[p,q]$ is the output of  Ko{\c{c}}'s, Ding's  or the proposed 2D NsDLCT:
\begin{align}\label{eq:Acc08}
G[p,q]={\cal O}_{\textmd{NsDLCT}}^{(\bb A, \bb B;\bb C,\bb D)}\left\{ {{g}[m,n]} \right\}.
\end{align}
In the
following simulations, 2D Hermite Gaussians (HGs) are used as the input.
The 1D HG of order $k$ is defined as
\begin{align}\label{eq:Acc12}
HG_{k}(x) = {\left( {\frac{1}{{{2^{k }}k!\sqrt{\pi} }}} \right)^{1/2}}{e^{ - \frac{x^2}{2}}}{H_k}(x),
\end{align}
where $H_k(x)$ is the $k$th-order physicists' Hermite polynomial.
The 2D HG of order $(k,l)$ is a separable function defined as
\begin{align}\label{eq:Acc16}
HG_{k,l}(x,y) = HG_{k}(x) HG_{l}(y).
\end{align}
Two signals composed of 2D HGs are
used:
\begin{align}\label{eq:Acc20}
g_1(x,y)&=HG_{1,2}(x,y)+HG_{3,1}(x,y),\nn\\
g_2(x,y)&=HG_{2,18}(x,y)+HG_{14,11}(x,y).
\end{align}
Fig.~\ref{fig:InputHG}(a) shows the $100\times100$ sampled $g_1(x,y)$, i.e. $g_1[m,n]$, with sampling intervals $\Delta_x=\Delta_y=0.25$,
while Fig.~\ref{fig:InputHG}(b) shows $165\times165$ $g_2[m,n]$ with $\Delta_x=\Delta_y=0.2$.
We can find out that $g_1[m,n]$ has energy more concentrated than $g_2[m,n]$.

\begin{figure}[t]
\centering
\includegraphics[width=.9\columnwidth,clip=true]{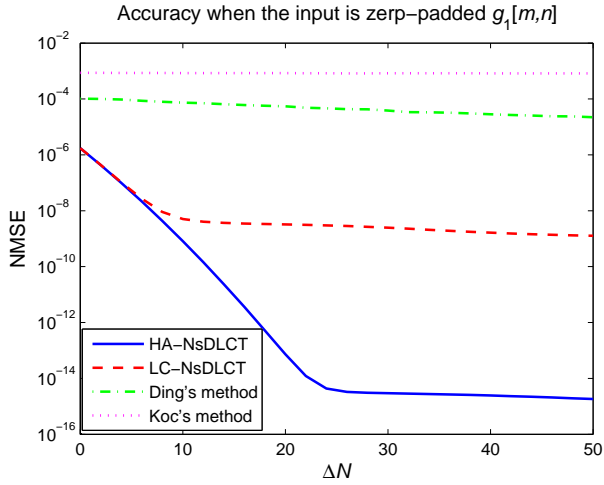}
\vspace*{-2pt}
\caption{Accuracy of Ko{\c{c}}'s method \cite{kocc2010fast},  Ding's  method \cite{ding2012improved}, the proposed 2D HA-NsDLCT and 2D LC-NsDLCT when $(\bb A_1, \bb B_1;\bb C_1,\bb D_1)$ in (\ref{eq:Acc24}) is used and $g_1[m,n]$ is zero-padded to $(100+\Delta N)\times(100+\Delta N)$.}
\label{fig:AccSmall}
\vspace*{-4pt}
\end{figure}

\begin{figure}[t]
\centering
\includegraphics[width=.9\columnwidth,clip=true]{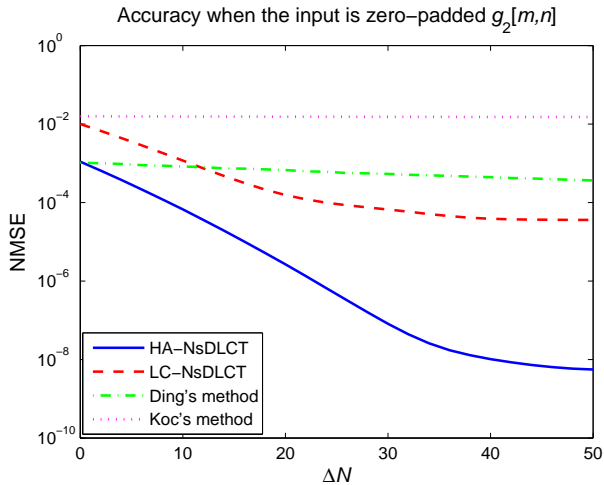}
\vspace*{-2pt}
\caption{Accuracy of Ko{\c{c}}'s method \cite{kocc2010fast},  Ding's  method \cite{ding2012improved}, the proposed 2D HA-NsDLCT and 2D LC-NsDLCT when $(\bb A_2, \bb B_2;\bb C_2,\bb D_2)$ in (\ref{eq:Acc28}) is used and $g_2[m,n]$ is zero-padded to $(165+\Delta N)\times(165+\Delta N)$.}
\label{fig:AccLarge}
\vspace*{-4pt}
\end{figure}

In order to analyze the accuracy by (\ref{eq:Acc04}), we need to derive the 2D NsLCTs of $g_1(x,y)$ and $g_2(x,y)$, denoted by $G_1(u,v)$ and $G_2(u,v)$, respectively.
However, there are no closed-form expressions for $G_1(u,v)$ and $G_2(u,v)$.
Therefore, we use the direct method in (\ref{eq:Intro32}) and discrete input with very large size and very small sampling intervals (i.e. $1024\times1024$  $g_1[m,n]$ and $g_2[m,n]$ with $\Delta_x=\Delta_y=0.078$) to approximate $G_1(p\Delta_u,q\Delta_v)$ and $G_2(p\Delta_u,q\Delta_v)$.
Fig.~\ref{fig:InputHG}(c) depicts the $100\times100$ approximate $G_1(p\Delta_u,q\Delta_v)$ with $\Delta_u=\Delta_v=0.25$ and ABCD matrix given by
\begin{align}\label{eq:Acc24}
\begin{bmatrix}
\bb A_1&\bb B_1\\
\bb C_1&\bb D_1
\end{bmatrix}=
\begin{bmatrix}
         0 &   1.1217 &  -0.7754 &  -0.3765\\
   -1.0934 &  -1.8826 &   1.1005 &   1.3878\\
    0.1697 &  -1.4013 &  -0.5352 &   1.2447\\
   -0.2014 &  -0.5209 &  -0.5916 &   0.3141
\end{bmatrix}.
\end{align}
Fig.~\ref{fig:InputHG}(d) shows the $165\times165$ approximate $G_2(p\Delta_u,q\Delta_v)$ with $\Delta_u=\Delta_v=0.2$ and ABCD matrix given by
\begin{align}\label{eq:Acc28}
\begin{bmatrix}
\bb A_2&\bb B_2\\
\bb C_2&\bb D_2
\end{bmatrix}=
\begin{bmatrix}
    0.3042 &  -0.2306 &   1.7626 &  -0.5090\\
   -0.2641 &  -0.7314 &  -1.2221 &  -1.2080\\
   -0.4765 &   0.4020 &  -0.1935 &  -0.0623\\
    0.3322 &   0.9671 &   0.7081 &   0.5295
\end{bmatrix}.
\end{align}

When the input is
$g_1[m,n]$, the NMSEs of Ko{\c{c}}'s method, Ding's method, the 2D HA-NsDLCT and 2D LC-NsDLCT are $8.7\times10^{-4}$, $1.0\times10^{-4}$, $1.7\times10^{-6}$ and  $1.7\times10^{-6}$, respectively.
The proposed 
methods have higher accuracy.
However, when the input is $g_2[m,n]$, the 2D HA-NsDLCT (NMSE $1.1\times10^{-3}$) and 2D LC-NsDLCT ($10^{-2}$) are better than  Ko{\c{c}}'s method ($1.6\times10^{-2}$) but
somewhat worse than Ding's method ($10^{-3}$).
This is because 
the output occupies larger space, as shown in Fig.~\ref{fig:InputHG}(d).
There would be some aliasing/overlapping effect around the boundary.
To solve this problem, the discrete input is first zero-padded to larger size, say $(N+\Delta N)\times(N+\Delta N)$. 
Fig.~\ref{fig:AccSmall} and Fig.~\ref{fig:AccLarge} show the accuracy
versus $0\leq\Delta N\leq50$ when the input is $g_1[m,n]$ and $g_2[m,n]$, respectively.
We can find out that the proposed 
methods significantly outperform Ko{\c{c}}'s and Ding's methods when $\Delta N$ is large enough.
Besides, these simulations verify the feature "high accuracy" of the 2D HA-NsDLCT.
The 2D HA-NsDLCT has higher accuracy than the 2D LC-NsDLCT.

\begin{figure}[t]
\centering
\includegraphics[width=1\columnwidth,clip=true]{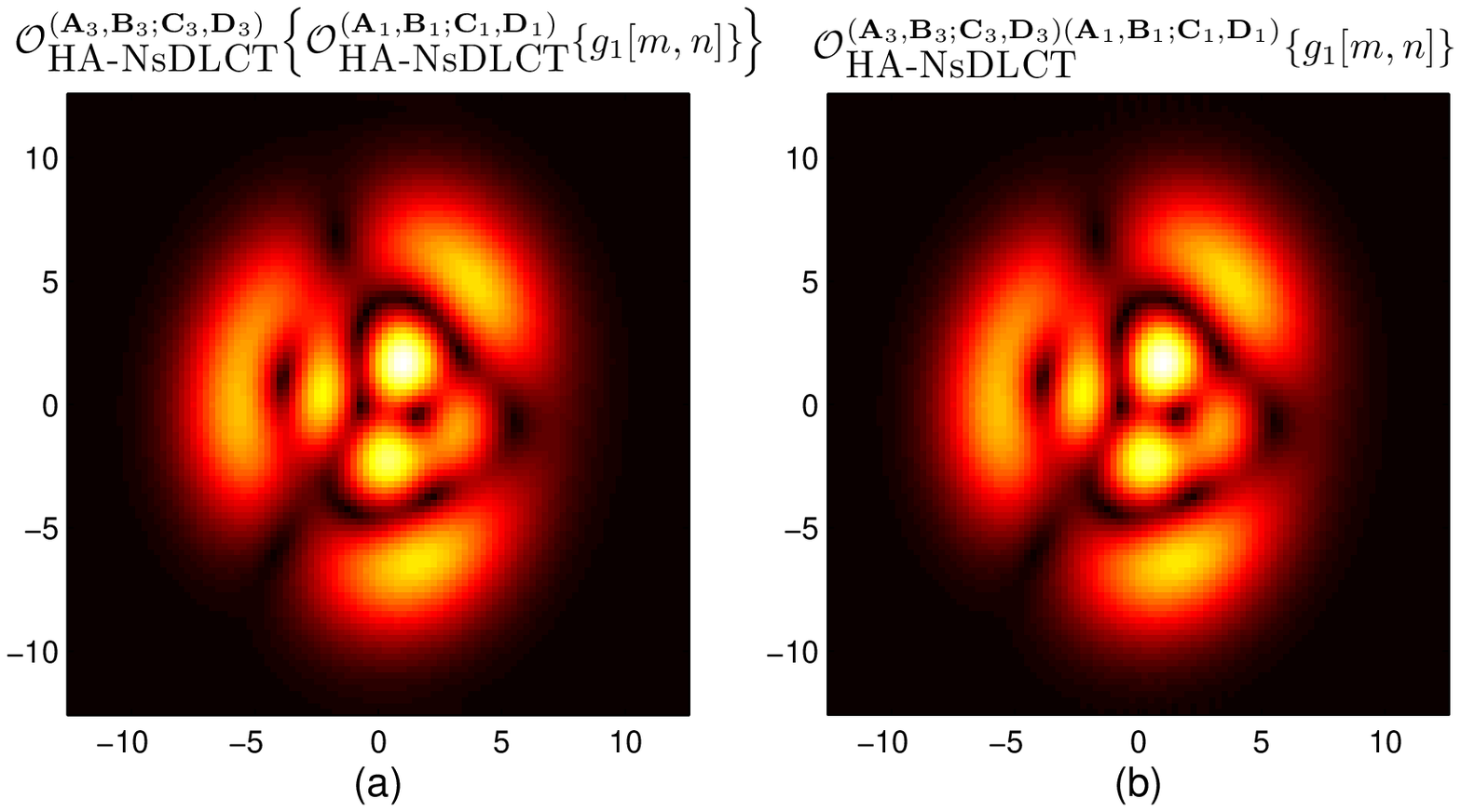}
\vspace*{-12pt}
\caption{
The 2D HA-NsDLCTs of $g_1[m,n]$ using (a) two steps of 2D HA-NsDLCTs ${\cal O}_{\textmd{NsDLCT}}^{(\bb A_3, \bb B_3;\bb C_3,\bb D_3)}
{\cal O}_{\textmd{NsDLCT}}^{(\bb A_1, \bb B_1;\bb C_1,\bb D_1)}$
and (b) one step of 2D HA-NsDLCT ${\cal O}_{\textmd{NsDLCT}}^{(\bb A_3, \bb B_3;\bb C_3,\bb D_3)(\bb A_1, \bb B_1;\bb C_1,\bb D_1)}$ supposing additivity property is satisfied,
where $(\bb A_1, \bb B_1;\bb C_1,\bb D_1)$ and $(\bb A_3, \bb B_3;\bb C_3,\bb D_3)$ are given in (\ref{eq:Acc24}) and (\ref{eq:Add24}), respectively.
The input is not zero-padded in these two examples.
}
\label{fig:AddSmall_1}
\vspace*{-4pt}
\end{figure}

\subsection{Additivity Property}\label{subsec:Add}
With the additivity property,
some
applications of 2D NsDLCT would
have lower computational complexity.
For example, consider that input $g[m,n]$ getting through two 
filters operating
in two different 2D NsLCT domains, say
\begin{align}\label{eq:Add04}
&g'[m,n]={\cal O}_{\textmd{NsDLCT}}^{(\bb A_1, \bb B_1;\bb C_1,\bb D_1)^{-1}}\left\{{\cal O}_{\textmd{NsDLCT}}^{(\bb A_3, \bb B_3;\bb C_3,\bb D_3)^{-1}}\left\{H_3[p_3,q_3] \frac{}{} \right.\right.\nn\\
&\ \ \ \left.\left.\cdot{\cal O}_{\textmd{NsDLCT}}^{(\bb A_3, \bb B_3;\bb C_3,\bb D_2)}\left\{H_1[p_1,q_1]{\cal O}_{\textmd{NsDLCT}}^{(\bb A_1, \bb B_1;\bb C_1,\bb D_1)}\left\{g[m,n]\right\}\right\}\right\}\right\},\!\!
\end{align}
where $H_1[p,q]$ and $H_3[p,q]$ are filters.
If the additivity property is satisfied, only three 2D NsDLCTs are required:
\begin{align}\label{eq:Add08}
&g'[m,n]={\cal O}_{\textmd{NsDLCT}}^{(\bb A_1, \bb B_1;\bb C_1,\bb D_1)^{-1}(\bb A_3, \bb B_3;\bb C_3,\bb D_3)^{-1}}\left\{H_3[p_3,q_3] \frac{}{} \right.\nn\\
&\ \ \ \ \left.\cdot{\cal O}_{\textmd{NsDLCT}}^{(\bb A_3, \bb B_3;\bb C_3,\bb D_2)}\left\{H_1[p_1,q_1]{\cal O}_{\textmd{NsDLCT}}^{(\bb A_1, \bb B_1;\bb C_1,\bb D_1)}\left\{g[m,n]\right\}\right\}\right\},\!\!
\end{align}
The existing 2D NsDLCTs don't satisfy the additivity property, and neither do the proposed methods.
However, we will show that the proposed 2D NsDLCTs have "approximate" additivity.

If a 2D NsDLCT has perfect additivity, it should satisfy
\begin{align}\label{eq:Add12}
{\cal O}_{\textmd{NsDLCT}}^{(\bb A_3, \bb B_3;\bb C_3,\bb D_3)}
{\cal O}_{\textmd{NsDLCT}}^{(\bb A_1, \bb B_1;\bb C_1,\bb D_1)} = {\cal O}_{\textmd{NsDLCT}}^{(\bb A_3, \bb B_3;\bb C_3,\bb D_3)(\bb A_1, \bb B_1;\bb C_1,\bb D_1)}.
\end{align}
Therefore, a 2D NsDLCT is referred to as being approximately additive
if the difference between the left side and right side of (\ref{eq:Add12}) is small enough.
The error of additivity is measured by the NMSE defined below:
\begin{align}\label{eq:Add16}
{\textmd{NMSE}}_{\textmd{add}} = \frac{\sum\limits_p^{}\sum\limits_q^{} {\left|G'[p,q]-G[p,q] \right|}^2 }{\sum\limits_p^{}\sum\limits_q^{} {\left|G[p,q] \right|}^2 },
\end{align}
where $G[p,q]$ and $G'[p,q]$ are given by
\begin{align}\label{eq:Add20}
G[p,q]&={\cal O}_{\textmd{NsDLCT}}^{(\bb A_3, \bb B_3;\bb C_3,\bb D_3)}\left\{
{\cal O}_{\textmd{NsDLCT}}^{(\bb A_1, \bb B_1;\bb C_1,\bb D_1)}\left\{g[m,n]\right\}\right\},\\
G'[p,q]&={\cal O}_{\textmd{NsDLCT}}^{(\bb A_3, \bb B_3;\bb C_3,\bb D_3)(\bb A_1, \bb B_1;\bb C_1,\bb D_1)}\left\{g[m,n]\right\},\label{eq:Add22}
\end{align}
respectively.

\begin{figure}[t]
\centering
\includegraphics[width=.9\columnwidth,clip=true]{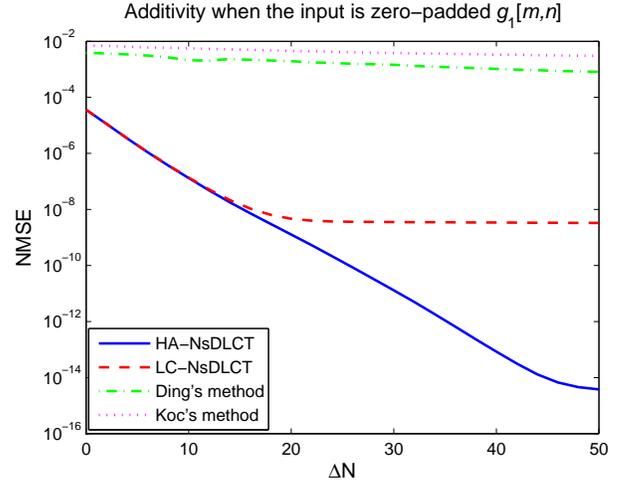}
\vspace*{-2pt}
\caption{Additivity of Ko{\c{c}}'s method \cite{kocc2010fast},  Ding's  method \cite{ding2012improved}, the proposed 2D HA-NsDLCT and 2D LC-NsDLCT when $(\bb A_1, \bb B_1;\bb C_1,\bb D_1)$ in (\ref{eq:Acc24}) and $(\bb A_3, \bb B_3;\bb C_3,\bb D_3)$ in (\ref{eq:Add24}) are used and $g_1[m,n]$ is zero-padded to $(100+\Delta N)\times(100+\Delta N)$.}
\label{fig:AddSmall_2}
\end{figure}

\begin{figure}[t]
\centering
\includegraphics[width=.9\columnwidth,clip=true]{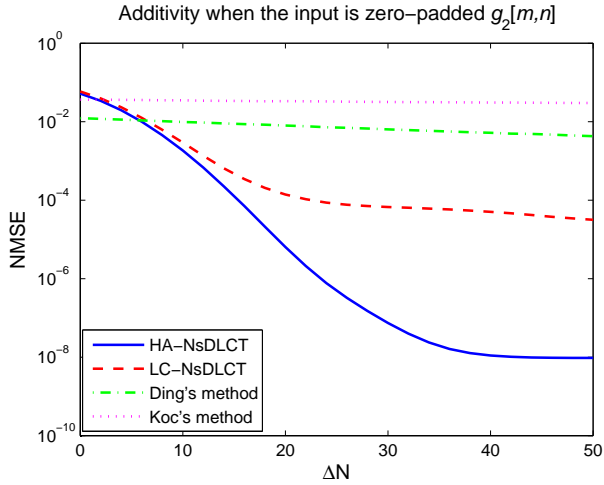}
\vspace*{-2pt}
\caption{Additivity of Ko{\c{c}}'s method \cite{kocc2010fast},  Ding's  method \cite{ding2012improved}, the proposed 2D HA-NsDLCT and 2D LC-NsDLCT when $(\bb A_2, \bb B_2;\bb C_2,\bb D_2)$ in (\ref{eq:Acc28}) and $(\bb A_4, \bb B_4;\bb C_4,\bb D_4)$ in (\ref{eq:Add28}) are used and $g_2[m,n]$ is zero-padded to $(165+\Delta N)\times(165+\Delta N)$.}
\label{fig:Addlarge_2}
\vspace*{-4pt}
\end{figure}

In the following, 
two simulations are presented.
The first one is using $(100+\Delta N)\times(100+\Delta N)$ zero-padded $g_1[m,n]$ as the input with $(\bb A_1, \bb B_1;\bb C_1,\bb D_1)$ given in (\ref{eq:Acc24}) and $(\bb A_3, \bb B_3;\bb C_3,\bb D_3)$ given by
\begin{align}\label{eq:Add24}
\begin{bmatrix}
\bb A_3&\bb B_3\\
\bb C_3&\bb D_3
\end{bmatrix}=
\begin{bmatrix}
   -0.4742  & -0.8700   & 2.4284 &  -2.6166\\
    4.1205  &  1.8038   & 2.6786 &  -7.5360\\
   -4.3025  & -0.6572   &-7.2020 &  12.8085\\
    3.8671  &  2.5257   &-0.6080 &  -2.8661
\end{bmatrix}.
\end{align}
Fig.~\ref{fig:AddSmall_1} shows the outputs of (\ref{eq:Add20}) and (\ref{eq:Add22}) when the original $g_1[m,n]$ (without zero-padding shown in Fig.~\ref{fig:InputHG}(a)) and 2D HA-NsDLCT are used.
Comparing the additivity by (\ref{eq:Add16}), the 2D  HA-NsDLCT and 2D LC-NsDLCT have NMSE $3.6\times10^{-5}$ lower than the Ko{\c{c}}'s method ($7.2\times10^{-3}$) and Ding's method ($3.9\times10^{-3}$).
If $g_1[m,n]$ is zero-padded to larger size, lower and more satisfactory NMSE can be achieved, as shown in Fig.~\ref{fig:AddSmall_2}.
This example shows that the proposed methods have approximate additivity.
\begin{figure}[t]
\centering
\includegraphics[width=1\columnwidth,clip=true]{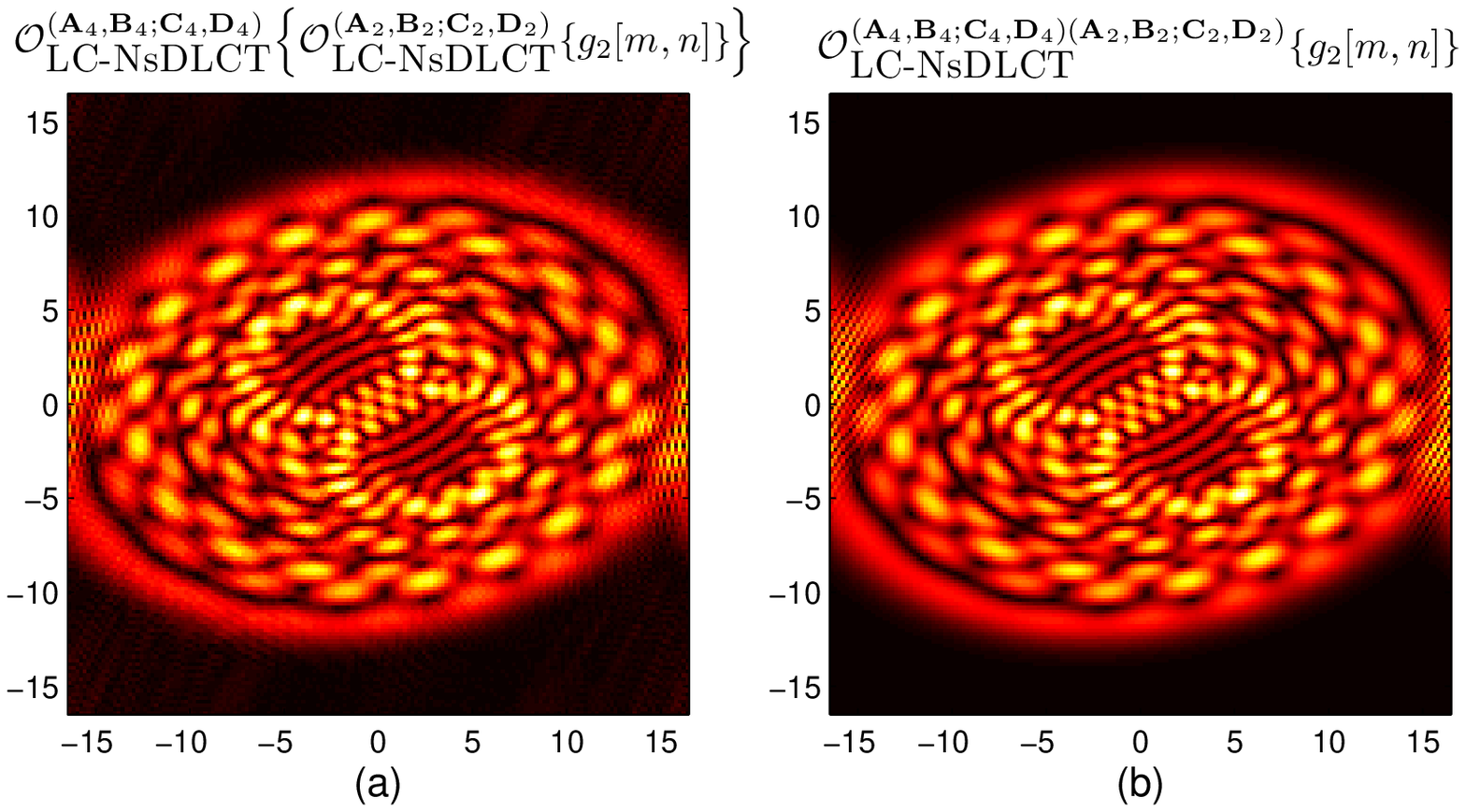}
\vspace*{-12pt}
\caption{
The 2D LC-NsDLCTs of $g_2[m,n]$ using (a) two steps of 2D LC-NsDLCTs ${\cal O}_{\textmd{NsDLCT}}^{(\bb A_4, \bb B_4;\bb C_4,\bb D_4)}
{\cal O}_{\textmd{NsDLCT}}^{(\bb A_2, \bb B_2;\bb C_2,\bb D_2)}$
and (b) one step of 2D LC-NsDLCT ${\cal O}_{\textmd{NsDLCT}}^{(\bb A_4, \bb B_4;\bb C_4,\bb D_4)(\bb A_2, \bb B_2;\bb C_2,\bb D_2)}$ supposing additivity property is satisfied,
where $(\bb A_2, \bb B_2;\bb C_2,\bb D_2)$ and $(\bb A_4, \bb B_4;\bb C_4,\bb D_4)$ are given in (\ref{eq:Acc28}) and (\ref{eq:Add28}), respectively.
The input is not zero-padded in these two examples.
}
\label{fig:Addlarge_1}
\end{figure}
In the second example, $(165+\Delta N)\times(165+\Delta N)$ zero-padded $g_2[m,n]$ gets through 2D NsDLCT with $(\bb A_2, \bb B_2;\bb C_2,\bb D_2)$ given in (\ref{eq:Acc28}) followed by 2D NsDLCT with $(\bb A_4, \bb B_4;\bb C_4,\bb D_4)$ shown below:
\begin{align}\label{eq:Add28}
\begin{bmatrix}
\bb A_4&\bb B_4\\
\bb C_4&\bb D_4
\end{bmatrix}=
\begin{bmatrix}
    0.7597  &  0.2418 &   1.4055 &   1.5125\\
    0.9305  &  0.1806 &   2.3170 &  -0.7412\\
   -0.0147  & -0.5068 &   0.5030 &  -0.7006\\
    0.4943  &  0.5059 &   1.8726 &   0.1543
\end{bmatrix}.
\end{align}
Fig.~\ref{fig:Addlarge_1} depicts the outputs of (\ref{eq:Add20}) and (\ref{eq:Add22}) when the original $g_2[m,n]$
(without zero-padding shown in Fig.~\ref{fig:InputHG}(b)) and 2D LC-NsDLCT are used.
These's some difference between these two outputs, especially around the boundary.
Analyzing the additivity by (\ref{eq:Add16}), the NMSEs of the 2D HA-NsDLCT and 2D LC-NsDLCT are $0.052$ and $0.059$, respectively, somewhat higher than the Ko{\c{c}}'s method ($0.037$) and Ding's method ($0.012$).
However, Fig.~\ref{fig:Addlarge_2} shows that the 2D HA-NsDLCT and 2D LC-NsDLCT have lower NMSEs than Ko{\c{c}}'s and Ding's methods and have approximate additivity when $g_2[m,n]$ is zero-padded to larger size.
The 2D HA-NsDLCT has approximate additivity with lower error than the 2D LC-NsDLCT because of its higher accuracy
feature.

\subsection{Reversibility Property}\label{subsec:Rev}
\begin{figure}[t]
\centering
\includegraphics[width=.9\columnwidth,clip=true]{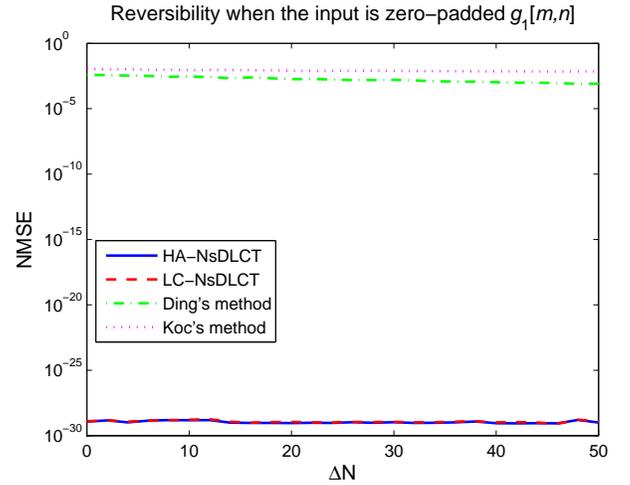}
\vspace*{-2pt}
\caption{Reversibility of Ko{\c{c}}'s method \cite{kocc2010fast},  Ding's  method \cite{ding2012improved}, the proposed 2D HA-NsDLCT and 2D LC-NsDLCT when $(\bb A_1, \bb B_1;\bb C_1,\bb D_1)$ in (\ref{eq:Acc24}) is used and $g_1[m,n]$ is zero-padded to $(100+\Delta N)\times(100+\Delta N)$.}
\label{fig:RevSmall}
\vspace*{-4pt}
\end{figure}

\begin{figure}[t]
\centering
\includegraphics[width=.9\columnwidth,clip=true]{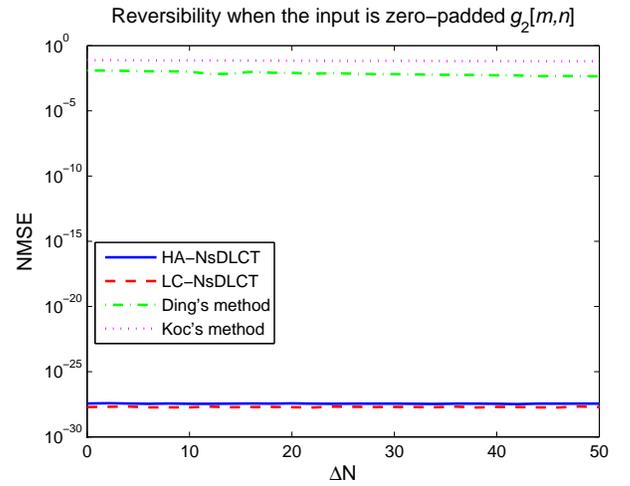}
\vspace*{-2pt}
\caption{Reversibility of Ko{\c{c}}'s method \cite{kocc2010fast},  Ding's  method \cite{ding2012improved}, the proposed 2D HA-NsDLCT and 2D LC-NsDLCT when $(\bb A_2, \bb B_2;\bb C_2,\bb D_2)$ in (\ref{eq:Acc28}) is used and $g_2[m,n]$ is zero-padded to $(165+\Delta N)\times(165+\Delta N)$.}
\label{fig:RevLarge}
\vspace*{-4pt}
\end{figure}

\begin{figure}[t]
\centering
\includegraphics[width=1\columnwidth,clip=true]{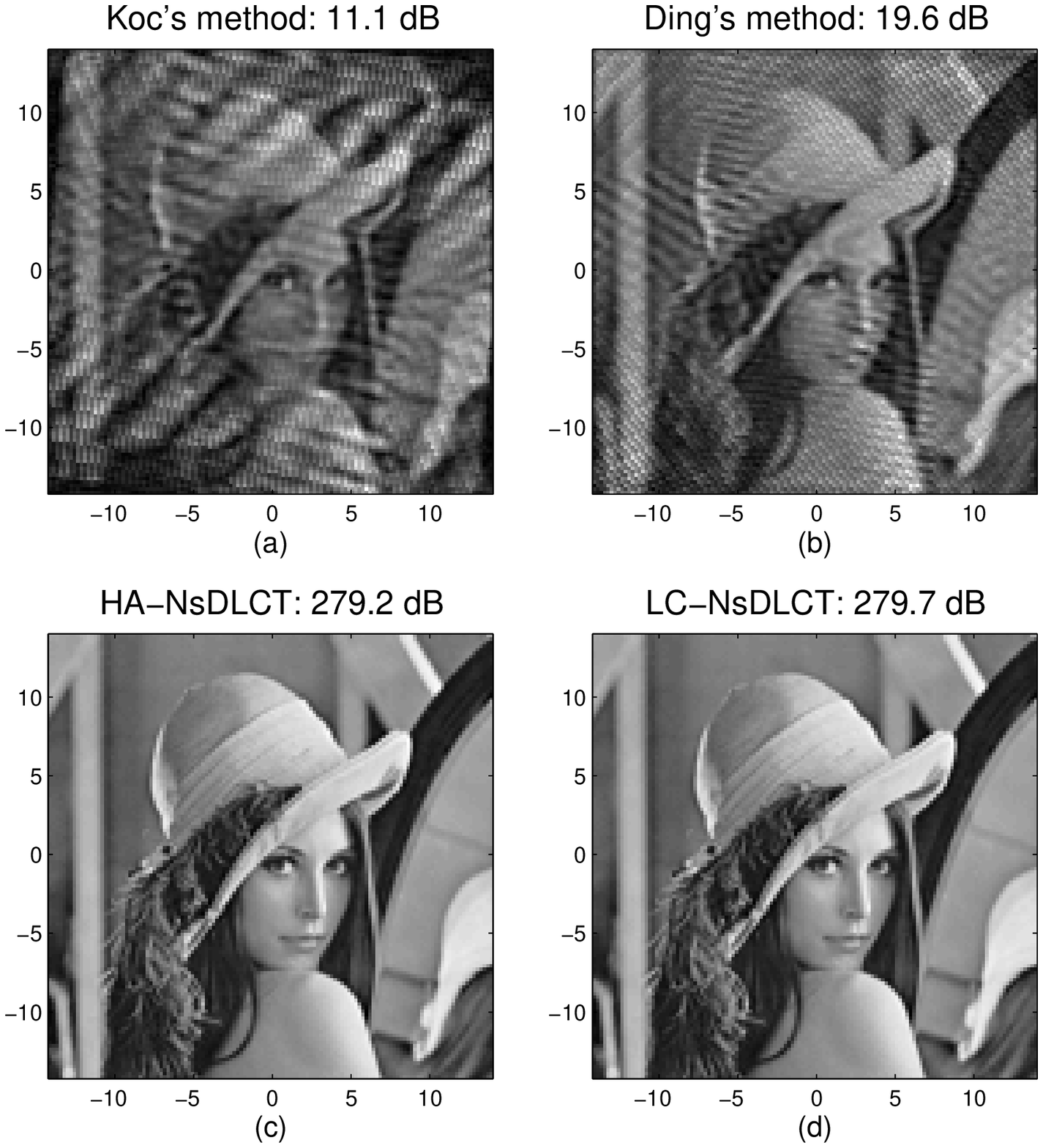}
\vspace*{-12pt}
\caption{
$128\times128$ reconstructed Lena image from performing  ${\cal O}_{\textmd{NsDLCT}}^{(\bb A_2, \bb B_2;\bb C_2,\bb D_2)^{-1}}$ on the 2D NsDLCT coefficients ${\cal O}_{\textmd{NsDLCT}}^{(\bb A_2, \bb B_2;\bb C_2,\bb D_2)}\left\{g[m,n]\right\}$: (a)Ko{\c{c}}'s method \cite{kocc2010fast}, (b)  Ding's  method \cite{ding2012improved}, (c) proposed 2D HA-NsDLCT and (d) proposed 2D LC-NsDLCT.
}
\label{fig:RevLena}
\vspace*{-4pt}
\end{figure}

The reversibility property is a special case of the additivity property.
A 2D NsDLCT with reversibility property should satisfy
\begin{align}\label{eq:Rev04}
{\cal O}_{\textmd{NsDLCT}}^{(\bb A, \bb B;\bb C,\bb D)^-1}\left\{
{\cal O}_{\textmd{NsDLCT}}^{(\bb A, \bb B;\bb C,\bb D)}\left\{g[m,n]\right\}\right\} = g[m,n].
\end{align}
Even without the additivity property, the reversibility property may be satisfied.
The reversibility is analyzed by the NMSE given by
\begin{align}\label{eq:Rev08}
{\textmd{NMSE}}_{\textmd{rev}} \!= \!\frac{\sum\limits_m^{}\sum\limits_n^{} {\left|{\cal O}_{\textmd{NsDLCT}}^{(\bb A, \bb B;\bb C,\bb D)^{-1}}\!\!\left\{
{\cal O}_{\textmd{NsDLCT}}^{(\bb A, \bb B;\bb C,\bb D)}\!\left\{g[m,n]\right\}\!\right\}\! \!- g[m,n] \right|}^2 }{\sum\limits_m^{}\sum\limits_n^{} {\left| g[m,n] \right|}^2 }.
\end{align}
When the input is zero-padded $g_1[m,n]$ and $(\bb A, \bb B;\bb C,\bb D)=(\bb A_1, \bb B_1;\bb C_1,\bb D_1)$, the NMSEs of the reversibility  are depicted in Fig.~\ref{fig:RevSmall},
while the NMSEs for zero-padded $g_2[m,n]$ and $(\bb A_2, \bb B_2;\bb C_2,\bb D_2)$ are shown in Fig.~\ref{fig:RevLarge}.
These examples verify that the proposed 2D NsDLCTs have perfect reversibility even 
without performing zero-padding (i.e. $\Delta N=0$).
At last, we use $128\times128$ Lena image as an example to present image reconstruction by each method.
The ABCD matrix $(\bb A_2, \bb B_2;\bb C_2,\bb D_2)$ in (\ref{eq:Acc28}) and $\Delta_x=\Delta_y=\Delta_u=\Delta_v=0.22$ is adopted.
The PSNRs of the 
reconstructed images
are shown in Fig.~\ref{fig:RevLena}.
Here, the zero-padding process is not employed.
The proposed 
methods have perfect reconstruction with PSNR about $279$ dB, while Ko{\c{c}}'s method and  Ding's  method have higher reconstruction errors, $11.1$ dB and $19.6$ dB, respectively.

Ding's  method would have perfect reversibility by using the sampling theorem and unitary discretization of 2D NsLCT in \cite{zhao2014two}.
However, there will be some restrictions on the locations of the output sampling points or the value of $\bb B$ in the ABCD matrix.
One can refer to \cite{zhao2014two} for more details and derivations of the restrictions.
Therefore, Ding's  method remains irreversible in most cases.

\section{Optical Applications}\label{sec:App}
The LCT can be used to model any lossless first-order optical system, also known as ABCD system \cite{bastiaans2007classification}, such as light propagation through free space, gradient index (GRIN) medium, elliptic GRIN medium, thin lens, spherical lens, cylindrical lens, tilted cylinder lens, prisms, Fourier transformer or a combination of several different spaces, media and lenses.
Thus, one can use the 2D NsDLCTs to simulate and analyze many 2D optical systems.

\subsection{Light Propagation Through Fourier Transformer and Elliptic GRIN Medium}\label{subsec:light}
If the refractive index distribution $n(x,y)$ of an elliptic GRIN medium satisfies
\begin{align}\label{eq:light04}
{n^2}(x,y) = n_0^2\left[ {1 - \frac{n_1}{n_0}{{(x + py)}^2} - \frac{n_2}{n_0}{{(qx + y)}^2}} \right],
\end{align}
where $n_0,n_1,n_2$ are the GRIN medium parameters,
the result of the light propagation in the medium can be expressed as \cite{yu1998fractional,pei2001two}
\begin{align}\label{eq:light08}
G_1(u,v)= {e^{ - j2\pi {n_0}\frac{L}{\lambda }}}{\cal O}_{\textmd{NsLCT}}^{(\bb A_1, \bb B_1;\bb C_1, \bb D_1)}\{g(x,y)\},
\end{align}
where $L$ and $\lambda$ are the propagation length and the wavelength, respective.
The ABCD matrix is given by
\begin{align}\label{eq:light12}
\begin{bmatrix}
{\bf{A}}_1&{\bf{B}}_1\\
{\bf{C}}_1&{\bf{D}}_1
\end{bmatrix}\! = \!\begin{bmatrix}
({\bf{S}}^T)^{-1}&{\bf{0}}\\
{\bf{0}}&{\bf{S}}^T
\end{bmatrix}\!\!
\begin{bmatrix}
{\bf{E}}&{\bf{F}}\\
{{\bf{F}}^{ - 1}}({{\bf{E}}^2} - I)&{\bf{F}}
\end{bmatrix}\!\!
\begin{bmatrix}
{\bf{S}}&{\bf{0}}\\
{\bf{0}}&{\bf{S}}^{-1}
\end{bmatrix},
\end{align}
where
\begin{align}\label{eq:light16}
\bb S=\begin{bmatrix}
1&p\\
q&1
\end{bmatrix},\
\bb E=\begin{bmatrix}
{\cos \alpha }&0\\
0&{\cos \beta }
\end{bmatrix},\
\bb F=\begin{bmatrix}
\sqrt{\frac{n_0}{n_1}}\frac{\sin \alpha }{k}&0\\
0&\sqrt{\frac{n_0}{n_2}}\frac{\sin \beta }{k}
\end{bmatrix},
\end{align}
and $k=\frac{2\pi}{\lambda}$, $\alpha=\frac{2L}{\pi}\sqrt{\frac{n_1}{n_0}}$ and $\beta=\frac{2L}{\pi}\sqrt{\frac{n_2}{n_0}}$.

In this section, we consider an optical system that consists of a Fourier transformer and an elliptic GRIN medium; that is,
the corresponding ABCD matrix is given by
\begin{align}\label{eq:light20}
\begin{bmatrix}
{\bf{A}}&{\bf{B}}\\
{\bf{C}}&{\bf{D}}
\end{bmatrix}= \begin{bmatrix}
{\bf{A}}_1&{\bf{B}}_1\\
{\bf{C}}_1&{\bf{D}}_1
\end{bmatrix} \begin{bmatrix}
{\bf{0}}&{\bf{I}}\\
-{\bf{I}}&{\bf{0}}
\end{bmatrix},
\end{align}
where $(\bb A_1, \bb B_1;\bb C_1, \bb D_1)$ is defined in (\ref{eq:light12}).

\begin{figure}[t]
\centering
\includegraphics[width=1\columnwidth,clip=true]{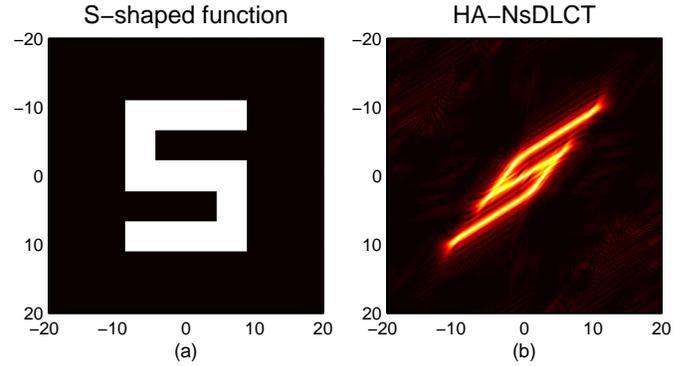}
\vspace*{-12pt}
\caption{
Optical system consisting of a Fourier transformer and an elliptic GRIN medium: (a) $257\times257$ sampled S-shaped function as the input and  (b) the output generated by the proposed 2D HA-NsDLCT with ABCD matrix given in (\ref{eq:light20}).
}
\label{fig:App1}
\vspace*{-4pt}
\end{figure}

A more accurate 2D NsDLCT is always a better choice to simulate the optical system.
Assume
\begin{align}\label{eq:light24}
&L=10^4\  \textmd{mm},\ \
\lambda=532\ \textmd{nm},\ \
p=0.6,\ \
q=0.2,\nn\\
&n_0=1.5,\ \
n_1=5\times10^{-8}\ \textmd{mm}^{-2},\ \
n_2=2\times10^{-8}\ \textmd{mm}^{-2},
\end{align}
and the input is a S-shaped function used in \cite{kocc2010fast}.
The $257\times257$ sampled S-shaped function is depicted in Fig.~\ref{fig:App1}(a), while the proposed 2D HA-NsDLCT without zero-padding the input is depicted in Fig.~\ref{fig:App1}(b).
The NMSE $0.038$ is lower than  Ko{\c{c}}'s method $0.24$ and Ding's method $0.073$.

\subsection{Self-Imaging Phenomena in Optics}\label{subsec:self}
For an optical system and its corresponding LCT, one can use the eigenfunction of the LCT to analyze the self-imaging phenomena of the optical system.
For 1D LCT, the eigenfunctions for all cases have been discussed in \cite{pei2002eigenfunctions}, and further simplified  into more compact closed forms without integral in \cite{pei2013differential}.
For the 2D NsLCT, the eigenfunctions for all cases of ABCD matrices  have been proposed in \cite{ding2011eigenfunctions}.
However, so far there is no general form for all cases of eigenfunctions, and the eigenfunctions in many cases are still expressed in integral form.
For example, consider an optical system corresponding to the following  ABCD matrix
\begin{align}\label{eq:self04}
\begin{bmatrix}
{\bf{A}}&{\bf{B}}\\
{\bf{C}}&{\bf{D}}
\end{bmatrix}= \begin{bmatrix}
   -0.1516  & -0.0982  & -1.5946  & -0.1626\\
   -0.0973   & 0.4641  &  0.0577  & -0.9005\\
    0.6387  &  0.0985  &  0.2636  & -0.0564\\
   -0.1039  &  0.9866  & -0.1599  &  0.1940
\end{bmatrix}.
\end{align}
In this case, the Method A in \cite{ding2011eigenfunctions} is used and results in the eigenfunctions of the following form:
\begin{align}\label{eq:self08}
\Psi (x,y)& = {e^{ - j0.392{x^2}}}\int\limits_{ - \infty }^\infty  {{e^{j\frac{{{{(x - \tau )}^2}}}{{{}1.3499}}}}} {\psi _1}( - 0.9368\tau  - 0.3601y)\nn\\
&\qquad\qquad\qquad\qquad\quad\cdot{\psi _2}(0.4388\tau  - 0.9027y)d\tau ,
\end{align}
where these parameters are obtained from a complicated procedure.
One can refer to \cite{ding2011eigenfunctions} for more details.
The functions $\psi_1$ and $\psi_2$ in (\ref{eq:self08}) are the eigenfunctions of two different 1D LCTs with parameters
$(0.1709,-0.8242;1.2515 , -0.1843)$ and $(0.5189,-0.8528;1.0116, 0.2646)$, respectively, in this case.
A different method instead of Method A would be used for a different ABCD matrix.
And in the discrete case, the samples of the eigenfunctions (\ref{eq:self08}) won't be orthogonal anymore.

\begin{figure}[t]
\centering
\includegraphics[width=1\columnwidth,clip=true]{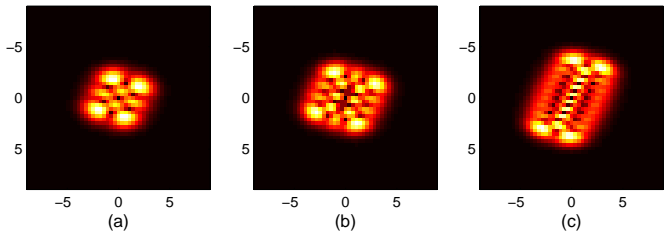}
\vspace*{-12pt}
\caption{
Self-imaging phenomena: three eigenvectors of the 2D HA-NsDLCT with ABCD matrix given in (\ref{eq:self04}).
}
\label{fig:App2}
\vspace*{-4pt}
\end{figure}

A general method that is suitable for all ABCD matrices and can generate a set of discrete orthogonal eigenfunctions is to calculate the eigenvectors of the 2D NsDLCT.
Given some $N\times N$ discrete input, reshape the input into an $N^2\times1$ column vector, say $\bb g$.
From (\ref{eq:HA28}), the proposed 2D HA-NsDLCT can be expressed as the following matrix form
\begin{align}\label{eq:self12}
{\bf{G}} = {\bf{L}} \cdot {\bf{g}} = {\bf{O}}_{{\rm{CM}}}^{({\bf{D'}} - {\bf{I}}){{{\bf{B'}}}^{ - {\bf{1}}}}}{{\bf{F}}^\dag }{\bf{O}}_{{\rm{CM}}}^{{\bf{B'}}}{\bf{F}}{\rm{ }}{\bf{O}}_{{\rm{CM}}}^{{{\bf{B}}^\prime }^{ - 1}({\bf{A}} - {\bf{I}})}{{\bf{F}}^\dag }{\bf{O}}_{{\rm{CM}}}^{\bf{H}}{\bf{F}}\ {\bf{g}}.
\end{align}
$\bb F$ is an $N^2\times N^2$ matrix obtained from the Kronecker product of two $N\times N$ DFT matrices.
$\bb F$ and $\bb F^\dag$ correspond to the 2D DFT and 2D IDFT, respectively.
The four ${\bf{O}}_{{\rm{CM}}}$ are $N^2\times N^2$ diagonal matrices performing 2D CMs.
Similarity, one can derive the matrix form for the 2D LC-NsDLCT.
Because the 2D HA-NsDLCT has perfect reversibility, the $N^2\times N^2$ matrix $\bb L$ in (\ref{eq:self12}) is unitary.
And one can obtain a set of orthogonal eigenvectors from $\bb L$, which can approximate the continuous eigenfunctions of the 2D NsLCT.
For example, when $N=51$, three eigenvectors of $\bb L$ with ABCD matrix given in (\ref{eq:self04}) are depicted in Fig.~\ref{fig:App2}.
The errors between the eigenvectors and the samples of continuous eigenfunctions are all below $10^{-4}$.

\section{Conclusions and Future Work}
To reduce computational complexity, the 2D NsLCT is decomposed into several simpler 2D operations.
Then, one can develop a low-complexity 2D NsDLCT by connecting the discrete versions of these 2D operations.
In this paper, a new decomposition called CM-CC-CM-CC decomposition is proposed.
Based on this decomposition, two types of 2D NsDLCTs are proposed, consisting of tow 2D discrete CMs and two 2D discrete CCs.
More precisely, only 2D discrete CMs, 2D DFTs and 2D IDFTs are used.
The proposed 2D NsDLCTs outperform the previous works in computational complexity, accuracy and additivity property.
A brief summary and comparisons are given in TABLE~\ref{tab:table2}.
Another type of decomposition called CC-CM-CC-CM decomposition is also proposed to ensure the reversibility of the proposed 2D NsDLCTs, so that the input signal/image can be perfectly reconstructed from the output of the proposed 2D NsDLCTs.

Additivity property is always a problem for discrete LCT, even for the 1D LCT \cite{pei2015fast}. Perfect additivity cannot be achieved usually because of the aliasing effect in spatial-frequency domain and overlapping in space domain. One straightforward method is zero-padding the input (as used in Sec. 5) to make the 2D NsDLCT approximate the 2D NsLCT, because the 2D NsLCT has perfect additivity. Another possible method is to restrict the ABCD matrix to some finite group, which may be developed in our future work.

\begin{table*}[t]
\small
\begin{center}
\setstretch{1.5}
\caption{Comparisons of the 2D NsDLCTs}\label{tab:table2}
\begin{tabular}{|l|c|c|c|c|}
\hline
 &Ko{\c{c}}'s method  \cite{kocc2010fast} & Ding's  method \cite{ding2012improved}& 2D HA-NsDLCT  & 2D LC-NsDLCT \\
\hline\hline
Complexity w.r.t. HA-NsDLCT & much higher& higher&reference& lower\\
\hline
Accuracy\textsuperscript{$*$} w.r.t. LC-NsDLCT &much lower& lower&higher& reference\\
\hline
Additivity\textsuperscript{$*$} (error w.r.t. LC-NsDLCT)  & approximate (much higher) & approximate (higher)& approximate (lower)& approximate (reference)\\
\hline
Reversibility & approximate & approximate& perfect & perfect\\
\hline
\multicolumn{5}{l}{\textsuperscript{$*$}{Suppose the input is first zero-padded to size large enough.}}
\end{tabular}
\end{center}
\vspace{-10pt}
\end{table*}

\end{document}